%% file: main.tex
\documentclass[10pt,twocolumn,letterpaper]{article}

\usepackage{cvpr}              %

\input{preamble}

\definecolor{cvprblue}{rgb}{0.21,0.49,0.74}
\usepackage[pagebackref,breaklinks,colorlinks,allcolors=cvprblue]{hyperref}

\def\paperID{} %
\def\confName{CVPR}
\def\confYear{2026}

\title{Learning Long-term Motion Embeddings for Efficient Kinematics Generation}

\author{
\begin{tabular}{c}
Nick Stracke$^{1,2,}$\footnotemark[1] \qquad
Kolja Bauer$^{1,2,}$\footnotemark[1] \qquad
Stefan Andreas Baumann$^{1,2}$ \\
Miguel Ángel Bautista$^{3}$ \qquad
Josh Susskind$^{3}$ \qquad
Bj\"orn Ommer$^{1,2}$\vspace{1mm} \\
{\normalsize $^1$ CompVis @ LMU \qquad $^2$ Munich Center for Machine Learning \qquad $^3$ Apple}\\
{\tt\small\rurl{compvis.github.io/long-term-motion}}
\end{tabular}
}

\begin{document}

\twocolumn[{%
    \maketitle
    \vspace{-11mm}
    \begin{center}
        \captionsetup{type=figure}{\includegraphics[width=0.9\textwidth]{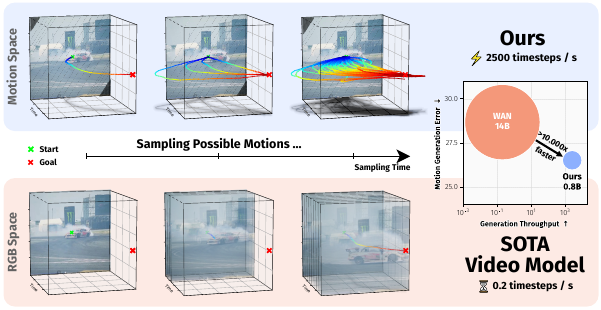}}
        \captionof{figure}{
       \textbf{Our approach enables extremely efficient, goal-conditioned kinematics generation and semantic motion reasoning.} We achieve this by learning a dense, temporally compressed motion space that allows goal-conditioned motion generation to be orders of magnitude faster than prior video models. While a video generative model has barely produced the first frame, our method can already generate multiple plausible motion trajectories connecting the start and goal, offering both speed and interpretability. }  
        \label{fig:teaser}
    \end{center}
}]

\def\thefootnote{}\footnotetext{\hspace{-14pt}$^*$ Equal Contribution}
\def\thefootnote{}\footnotetext{\hspace{-14pt}Experimentation, including use of pre-trained models, was completed by university collaborators}

\input{sec/0_abstract}

\input{sec/1_intro}

\input{sec/2_related}

\input{sec/3_method}

\input{sec/4_exps}

\input{sec/5_conclusion}

\section*{Acknowledgments}
This project has been supported by \todo{Apple}, the Horizon Europe project ELLIOT (GA No.\ 101214398), the project ``GeniusRobot'' (01IS24083) funded by the Federal Ministry of Research, Technology and Space (BMFTR), the BMWE ZIM-project (No.\ KK5785001LO4) ``conIDitional LoRA'', the German Federal Ministry for Economic Affairs and Energy within the project ``NXT GEN AI METHODS - Generative Methoden für Perzeption, Prädiktion und Planung'', and the bidt project KLIMA-MEMES. The authors gratefully acknowledge the Gauss Center for Supercomputing for providing compute through the NIC on JUWELS/JUPITER at JSC and the HPC resources supplied by the NHR@FAU Erlangen.

Further, we would like to thank Owen Vincent for continuous technical support.

{
    \small
    \bibliographystyle{ieeenat_fullname}
    \bibliography{main}
}

\input{sec/X_suppl}

\end{document}

%% file: preamble.tex
\usepackage{adjustbox}
\usepackage{graphicx}
\usepackage{caption}
\usepackage{subcaption}
\usepackage{printlen}
\usepackage{float}
\usepackage{wrapfig}
\usepackage{multirow}
\usepackage{booktabs}
\usepackage{pifont}
\usepackage{makecell}
\usepackage{nicefrac}
\usepackage{url}
\usepackage{yhmath}
\usepackage{stmaryrd}
\usepackage{verbatim}
\usepackage{amsmath}
\usepackage{algorithm}
\usepackage{algpseudocode}
\usepackage{amsfonts}
\usepackage{microtype}
\usepackage{pgf}
\usepackage{pgfplots}
\usepackage{tabularx}
\usepackage{mathtools}
\usepackage{ifthen}
\usepackage{soul}
\usepackage{tikz}
\usetikzlibrary{angles,quotes,3d,math,arrows.meta,calc,positioning,fit,backgrounds,decorations.pathreplacing,calligraphy,shapes,shapes.multipart}
\usepackage[table]{xcolor}
\usepackage[accsupp]{axessibility}

\newcommand{\todo}[1]{{\color{red}#1}}
\newcommand{\TODO}[1]{\textbf{\color{red}[TODO: #1]}}

\newcommand{\nick}[1]{\textbf{{\color{ourturquoise}[Nick: #1]}}}
\newcommand{\miguel}[1]{\textbf{{\color{ourblue}[Miguel: #1]}}}
\newcommand{\stefan}[1]{\textbf{{\color{ouryellow}[Stefan: #1]}}}

\newcommand{\omitforsubmission}[1]{{\color{ourpurple}#1}}
\newcommand{\remove}[1]{{\color{ourorange}#1}}
\newcommand{\draft}[1]{{\color{gray}#1}}
\newcommand{\compactversion}[1]{#1}

\renewcommand{\TODO}[1]{}
\renewcommand{\todo}[1]{#1}
\renewcommand{\nick}[1]{}
\renewcommand{\miguel}[1]{}
\renewcommand{\stefan}[1]{}
\renewcommand{\compactversion}[1]{}
\renewcommand{\omitforsubmission}[1]{}
\renewcommand{\draft}[1]{}
\renewcommand{\remove}[1]{}

\newcommand\rurl[1]{%
  \href{https://#1}{\nolinkurl{#1}}%
}

\definecolor{ourgreen}{RGB}{46, 204, 113}
\definecolor{ourgreenborder}{RGB}{39, 174, 96}
\definecolor{ourblue}{RGB}{52, 152, 219}
\definecolor{ourblueborder}{RGB}{41, 128, 185}
\definecolor{ourorange}{RGB}{230, 126, 34}
\definecolor{ourorangeborder}{RGB}{211, 84, 0}
\definecolor{ourred}{RGB}{231, 76, 60}
\definecolor{ourredborder}{RGB}{192, 57, 43}
\definecolor{ouryellow}{RGB}{241, 196, 15}
\definecolor{ouryellowborder}{RGB}{243, 156, 18}
\definecolor{ourpurple}{RGB}{155, 89, 182}
\definecolor{ourpurpleborder}{RGB}{142, 68, 173}
\definecolor{ourturquoise}{RGB}{26, 188, 156}
\definecolor{ourturquoiseborder}{RGB}{22, 160, 133}
\definecolor{ourturquoise}{RGB}{26, 188, 156}
\definecolor{ourturquoiseborder}{RGB}{22, 160, 133}
\definecolor{ourwhite}{RGB}{236, 240, 241}
\definecolor{ourwhiteborder}{RGB}{189, 195, 199}
\definecolor{ourgray}{RGB}{149, 165, 166}
\definecolor{ourgrayborder}{RGB}{127, 140, 141}

\usepackage[table]{xcolor}

%% file: sec/0_abstract.tex
\begin{abstract}
Understanding and predicting motion is a fundamental component of visual intelligence. Although modern video models exhibit strong comprehension of scene dynamics, exploring multiple possible futures through full video synthesis remains prohibitively inefficient. 
We model scene dynamics orders of magnitude more efficiently by directly operating on a \textbf{long-term motion embedding} that is learned from large-scale trajectories obtained from tracker models.
This enables efficient generation of long, realistic motions that fulfill goals specified via text prompts or spatial pokes. To achieve this, we first learn a highly compressed motion embedding with a temporal compression factor of $64\times$. In this space, we train a conditional flow-matching model to generate motion latents conditioned on task descriptions. The resulting motion distributions outperform those of both state-of-the-art video models and specialized task-specific approaches.
\end{abstract}

%% file: sec/1_intro.tex
\vspace{-2mm}
\section{Introduction}
\label{sec:intro}

Understanding and predicting motion is a fundamental trait of visual intelligence, yet current learning-based approaches with the goal of understanding and generating motion either focus on low-level motion information (optical flow or spatially sparse tracks) or conflate motion with appearance in video generative models, ultimately having to model per-pixel changes over time and requiring large amounts of compute to model this high-dimensional signal. In this paper, we propose to address these limitations by learning a long-term \textit{motion embedding}: a compact, semantic latent representation that aggregates global kinematic structure, integrates information across trajectories, and models how motion evolves over extended time horizons. Unlike flow, which describes only instantaneous displacements, or tracks, which record sparse point correspondences, our embedding learns a continuous, scene-level representation of motion dynamics that generalizes beyond the observed samples and supports reasoning and generation in a unified latent space.

Our approach is rooted in two insights.  
First, learning useful motion representations requires more than tracking what moves: it demands reasoning about how things \textit{can} move, how motions aggregate across objects, and forecasting what complex future behaviors are plausible in a given scene.  
Second, while direct modeling of video~\cite{wan2025wan,Veo3,yang2024cogvideox} or flow~\cite{shi2024motioni2v,niu2024mofa} provides access to rich visual signals, these representations are high-dimensional and computationally intensive, and they cannot be strongly temporally compressed without significant information loss.
In contrast, trajectories (tracks) are far more compact and interpretable, but lack the ability to generalize or aggregate motion contextually, and remain limited to the specific points sampled by the tracker.

To address these limitations, we introduce a two-stage framework that learns and predicts in a motion space for long-term kinematics.  
As the first step, we train a model to map sets of sparse tracker-derived motion samples into a continuous latent embedding.  
Unlike raw tracks, our latent motion space can be queried at any spatial position, enabling dense, context-aware motion predictions that go far beyond the tracker inputs.  
Secondly, we leverage the learned motion space for higher-order generative reasoning: a conditional flow matching model operates directly in the motion space, generating plausible, goal-directed motion under complex prompts such as text or spatial queries.  

Our motion space thus sits at a new level of abstraction: longer than flow, richer than tracks, and far more efficient than video.  
It aggregates global kinematic context, supports reasoning about motion beyond observed samples, and provides a standardized interface for downstream models.  %
By focusing on kinematics, i.e., how things \textit{can} move, not just what happens frame-to-frame, we enable new forms of motion reasoning and generative modeling not possible with prior representations.

We demonstrate that our approach delivers substantial efficiency and reasoning gains across diverse motion tasks, outperforming raw tracks and enabling controllable motion generation with orders of magnitude greater compression than video models.  
Our framework thus establishes motion spaces as a powerful substrate for both large-scale motion understanding and semantic kinematics generation.

\todo{
\paragraph{Contributions:}
\begin{enumerate}
    \item We propose a long-term motion embedding that captures the essential motion patterns of a scene from a single image, avoiding the overhead of full video generation.
    \item We develop a goal-conditioned motion generator trained via flow matching on open-set data for efficient generative motion reasoning, operating entirely on the learned motion embedding.
    \item We show that strong temporal compression improves learning, yielding better motion quality, more coherent dynamics, and faster training and inference.
    \item 
    Through extensive evaluations on open-set internet videos and robotics benchmarks, we show that our approach outperforms specialized trajectory predictors and video-based baselines while being substantially more efficient.
\end{enumerate}
}

%% file: sec/2_related.tex
\section{Related Work}
\begin{figure*}[t]
    \centering
    \includegraphics[width=0.8\linewidth]{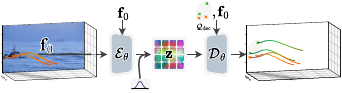}
    \caption{
    \textbf{Our approach to learn a dense motion space.} Sparse tracker trajectories and the start frame are encoded into a latent motion grid, which enables dense reconstruction at arbitrary spatial query points. The model jointly attends over trajectory tokens and frame features, producing temporally consistent, spatially dense motion predictions.
    }
    \label{fig:s1_overview}
\end{figure*}
\paragraph{Motion Representations}
Learning compact representations of motion has been studied from several perspectives, ranging from classical trajectory modeling to modern autoencoder-based approaches. 
Early works such as \emph{An Uncertain Future}~\cite{walker2016uncertain} explored stochastic VAEs for predicting future motion conditioned on a single image. 
More recent efforts adopt masked autoencoding or vector quantization to reconstruct trajectories or motion fields. 
Examples include trajectory autoencoders for anomaly detection or policy learning~\cite{allen2025trajan, collins2025amplify}, and two-stage models such as WHN~\cite{boduljak2025happens} that encode dense grid trajectories and generate in latent space without temporal compression. 

In contrast, our goal is to learn a \textit{semantic and structured motion space} that captures generic kinematics to enable open-domain planning. 

\paragraph{Motion in Video Models}
Most modern generative video models implicitly learn motion alongside appearance synthesis. 
Large-scale diffusion architectures such as Wan~\cite{wan2025wan}, Veo 3~\cite{Veo3} and others~\cite{yang2024cogvideox,HaCohen2024LTXVideo,Sora2,blattman2023svd} predict future frames without an explicit notion of motion. While such models excel at visual fidelity, the learned motion is entangled with texture and lighting, making kinematic reasoning and control difficult. Moreover, latent video autoencoders typically employ modest compression ratios (typically 4$\times$–8$\times$), as stronger compression severely degrades visual detail. These modest compression ratios imply that large compute budgets are often needed to learn useful video generative models.
Recent works have also explored world modeling in pre-trained feature spaces rather than RGB pixels, for example, by predicting the temporal evolution of DINO~\cite{oquab2023dinov2} features~\cite{zhou2024dino,baldassarre2025back,karypidis2024dino}. 
These approaches demonstrate that dynamics can be learned efficiently in a latent representation space, yet they do not provide an explicit notion of motion or kinematic structure. 
Their dense feature dynamics remain entangled with appearance and lack the interpretability and controllability of explicit trajectory-based motion representations. 
In contrast, our method models motion directly through sparse trajectories in a dedicated latent space, yielding a compact and semantically structured representation of kinematics.

\paragraph{Motion-conditioned Video Generation}
A separate line of work explicitly models motion as an intermediate representation before video synthesis. 
Motion I2V~\cite{shi2024motioni2v}, MoVideo~\cite{liang2024movideo}, and MOFA~\cite{niu2024mofa} first generates optical flow or depth maps that guide subsequent frame generation. 
VideoJam~\cite{chefer2025videojam} combines motion and appearance branches through joint diffusion, while dragging-based methods~\cite{shin2024instantdrag,shi2024dragdiffusion,wu2024draganything, geng2025motion, blattmann2021ipoke} manipulate motion vectors interactively to steer local frame changes, provide intuitive spatial or semantic control over motion dynamics. 
Although these models explicitly handle motion, it remains defined at the pixel level and tied to visual rendering. 
In contrast, our framework models motion itself as the generative domain by learning a dedicated latent motion space, which can later drive downstream synthesis or reasoning tasks without relying on pixel data.

\paragraph{Goal-Conditioned Motion Generation}
Controllable motion generation has received increasing attention in both vision and robotics. 
Trajectory-based models such as ATM~\cite{wen2023any}, Traj-MoE~\cite{yang2025tra}, Track2Act~\cite{bharadhwaj2024track2act}, and Amplify~\cite{collins2025amplify} predict trajectories conditioned on high-level cues like start frames or language, often for action policy learning. \cite{thakkar2026forecasting} uses a diffusion model to predict trajectories for animals based on a start frame and a brief motion history. 
Vision-Language-Action (VLA)~\cite{lee2025molmoact,zhang2025dreamvla,zheng2025tracevla} use trajectory prediction as a means of grounding vision-language models in physical actions. 
Our method builds on this idea of controllable generation but operates fully within the latent motion space learned by our autoencoder. 
The second stage employs a conditional flow matching model that generates motion latents conditioned on text or spatial pokes, enabling efficient and interpretable motion synthesis. 
Unlike prior approaches that predict explicit trajectories for downstream control or learn motion implicitly through appearance, our model unifies both perspectives by learning a generative motion prior in latent space that is compact and semantically grounded.

%% file: sec/3_method.tex
\section{Method}

We aim to first learn a compact and dense latent representation of trajectories obtained from off-the-shelf trackers that we can later use to model and manipulate kinematics. 

We represent a track as a sequence of $x, y$ coordinates in a normalized grid, i.e. $x,y \in [-1,1]$. We denote a single track as $\mathbf{x}_i = ([x_0, y_0], ..., [x_t, y_t], ..., [x_{T-1}, y_{T-1}])$, with $t$ being the timestep, representing the motion of a tracked point across time within a normalized image coordinate system.

\begin{figure*}
    \centering
    \includegraphics[width=0.8\linewidth]{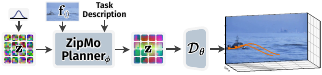}
    \caption{\textbf{Model architecture to generate in learned motion space.} We train a conditional flow matching model that learns a vector field over latent motion grids. We condition on either pokes~\cite{blattmann2021ipoke} or text prompts, enabling controllable and semantically coherent motion synthesis in the learned motion space. The frame $\mathbf{f}_0$ provides context over the scene.
    }
    \label{fig:s2-overview}
\end{figure*}

\subsection{Learning Motion Spaces}
\label{subsec:learning_motion_spaces}

\paragraph{Overview} To compress trajectories into a compact latent representation, we train a variational autoencoder~\cite{kingma2013vae} $\mathcal{F}_\theta = (\mathcal{E}_\theta,\, \mathcal{D}_\theta)$ that maps sets of sparse, partially masked trajectories to a latent grid and reconstructs them from it. 
The encoder $\mathcal{E}_\theta$ processes a set of trajectories $\mathbf{X} = \{\mathbf{x}_0, \dots, \mathbf{x}_{N-1}\}$, where each trajectory $\mathbf{x}_i = (\mathbf{x}_{i,0}, \dots, \mathbf{x}_{i,T-1})$ contains spatial coordinates $\mathbf{x}_{i,t} = [x_{i,t}, y_{i,t}]$, together with an embedding of the first video frame $\mathbf{f_0}$ (e.g., a DINO~\cite{zhou2024dino} feature map). 
The encoder produces a latent representation
\begin{equation}
    \mathcal{E}_\theta: \bigl(\underbrace{\bigl\{(\mathbf{x}_{i,t})_{t \in \mathbf{t}}\bigr\}_{i\in \mathcal{I}}}_{\text{trajectories}}, \underbrace{\phantom{\bigl\{}\mathbf{f}_0\phantom{\bigr\}}}_\text{frame}\bigr) \mapsto \underbrace{\phantom{\bigl\{}\mathbf{z}\phantom{\bigr\}}}_\text{latent},
\end{equation}
with $\mathbf{z} \in \mathbb{R}^{H\times W \times D}$ where $H, W$ denote the spatial grid size and $D$ the number of latent channels. %

The latent grid acts as a compact \textit{motion space}, summarizing the kinematic content of all trajectories in a video. 
The decoder $\mathcal{D}_\theta$ reconstructs future motion at arbitrary spatial locations and time indices through a set of query tokens,
\begin{equation}
    \mathcal{D}_\theta: \bigl(\underbrace{\bigl\{\mathbf{x}_{j,0}\bigr\}_{j\in \mathcal{J}}}_{\!\!\!\!\!\!\text{query points } \mathcal{Q}_\text{dec}\!\!\!\!\!\!}, \mathbf{z}, \mathbf{f}_0\bigr) \mapsto \underbrace{\bigl\{(\hat{\mathbf{x}}_{j,t})_{t\in \mathbf{t}}\bigr\}_{j\in \mathcal{J}}}_\text{trajectories},
\end{equation}
where $\mathcal{Q}_\text{dec} = \{\mathbf{x}_{j,0}\}_{j\in \mathcal{J}}$ denotes query points in normalized coordinates. 
Importantly, query points are not restricted to those used during encoding. 
While the autoencoder is trained with sparse trajectories, the learned latent representation allows decoding motion at any position on the start frame, effectively enabling dense motion reconstruction. 

\paragraph{Encoder} Each individual trajectory sample $\mathbf{x}_{i,t}$ is Fourier-embedded using random frequencies drawn from a Gaussian distribution~\cite{tancik2020fourier}. 
Temporal and trajectory identity information are encoded using a 3D rotary positional embedding (RoPE)~\cite{su2024roformer}. It jointly encodes each sample’s time index $t$ and the start position $[x_{i,0}, y_{i,0}]$, which anchors the trajectory’s identity. Token values and positional embeddings are thus computed as follows: 
\begin{align}
    \!\!\!\!\!\!\mathrm{tok}(\mathbf{x}_{i,t}) &= \mathrm{MLP}([\mathcal{F}(x_t) \mid \mathcal{F}(y_t)]) &&{\color{ourgrayborder}\!\!\triangleright\ \text{Encoding}}\!\!\!\!\label{eq:enc}\\
    \!\!\!\!\!\!\mathrm{PE}(\mathbf{x}_{i,t}) &= [\overbracket{\underbrace{\mathbf{R}(x_0)}_{d_k/4}\mid\underbrace{\mathbf{R}(y_0)}_{d_k/4}}^\text{start position}\mid\overbracket{\underbrace{\mathbf{R}(t)}_{d_k/4}}^\text{time}\mid\overbracket{\underbrace{\phantom{(}\mathbf{1}\phantom{)}}_{d_k/4}}^\text{empty}] &&{\color{ourgrayborder}\!\!\triangleright\ \text{RoPE}}\label{eq:rope}\!\!\!\!%
\end{align}
The latent grid $\textbf{z}$ is initialized as a learnable embedding broadcasted to the desired latent grid size $H \times W$. We differentiate between individual tokens of the latent grid using RoPE:
\begin{align}
    \mathrm{PE}(\mathbf{z}) &= [\overbracket{\underbrace{\mathbf{R}(h)}_{d_k/4}\mid\underbrace{\mathbf{R}(w)}_{d_k/4}}^\text{spatial position}\mid\overbracket{\underbrace{\phantom{(}\mathbf{\hspace{12pt}1\hspace{12pt}}\phantom{)}}_{d_k/2}}^\text{empty}]\hspace{8pt}{\color{ourgrayborder}\triangleright\ \text{RoPE}}\!\! 
\end{align}
We follow~\cite{crowson2024hdit,barbero2025round} and only apply RoPE partially to encourage the model to rely more on semantic information instead of only positional information. In total, this yields $N\cdot T$ trajectory tokens jointly processed with the tokens of the latent grid $\textbf{z}$. 
All tokens interact through global self-attention~\cite{vaswani2017attention} with interleaved cross-attention to the start-frame features $\mathbf{f_0}$. 
The encoder outputs the mean and log-variance of a Gaussian posterior
\begin{equation}
    q_\theta(\mathbf{z} \mid \mathbf{X}, \mathbf{f_0}) = 
    \mathcal{N}\!\left(\boldsymbol{\mu}_\theta(\mathbf{X}, \mathbf{f_0}),
    \, \mathrm{diag}\big(\boldsymbol{\sigma}^2_\theta(\mathbf{X}, \mathbf{f_0})\big)\right).
\end{equation}
\paragraph{Decoder} The decoder follows a masked-autoencoder (MAE)~\cite{he2022masked} design. 
Each query token encodes its temporal index $t_q$ and start position $(x_{q,0}, y_{q,0})$ using RoPE similar to the encoder, and attends to the latent grid $\mathbf{z}$ and the start-frame features $\mathbf{f_0}$ via cross-attention. 
The decoded query tokens are projected to $(x,y)$ coordinates through a small MLP, yielding motion predictions at the specified query positions.

Training follows a $\beta$-VAE objective~\cite{higgins2017betavae} combining an L1 reconstruction loss and a KL regularization term:

\begin{align}
    \mathcal{L} = &\underbrace{\frac{1}{|\mathcal{I}|}\sum_{i\in \mathcal{I}} \bigl\|\mathcal{D}_\theta\bigl(\mathbf{x}_{i,0}; \mathbf{z}, \mathbf{f}_0\bigr) - \mathbf{x}_{i}\bigr\|_1}_{\text{reconstruction}}\\
    &+ \underbrace{\frac{1}{|\mathcal{J}_\mathrm{mae}|}\sum_{j\in \mathcal{J}_\mathrm{mae}}\bigl\|\mathcal{D}_\theta\bigl(\mathbf{x}_{j,0}; \mathbf{z}, \mathbf{f}_0\bigr) - \mathbf{x}_{j}\bigr\|_1}_{\text{masked reconstruction: }\mathcal{J}_\mathrm{mae}\ \cap\ \mathcal{I} = \varnothing}\\
    &+ \beta \underbrace{D_{KL}[q_\theta(\mathbf{z}\mid \bigl\{(\mathbf{x}_{i,t})_{t \in \mathbf{t}}\bigr\}_{i\in \mathcal{I}}, \mathbf{f}_0\bigr)\parallel p(\mathbf{z})]}_\text{latent regularization}
\end{align}

where $\{\mathbf{x}_{i,0}\}_{i \in \mathcal{I}}$ are the query points corresponding to encoded trajectories (autoencoding loss) and $\{\mathbf{x}_{j,0}\}_{j \in \mathcal{J}_\mathrm{mae}}$ are randomly held-out points not seen by the encoder (masked reconstruction loss). $\beta$ controls the strength of the KL term.

This stage learns a semantic and structured motion bottleneck. 
The latent grid $\mathbf{z}$ captures the essential kinematics while enabling motion reconstruction at arbitrary spatial locations. 
Because trajectories are inherently low-dimensional and disentangled from appearance, the encoder achieves strong temporal compression without sacrificing semantic coherence. 
The resulting latent space provides a compact and interpretable \textit{motion space}, forming the foundation for the subsequent flow-matching generation stage.

\begin{figure*}[t]
    \centering
    \includegraphics[width=\linewidth]{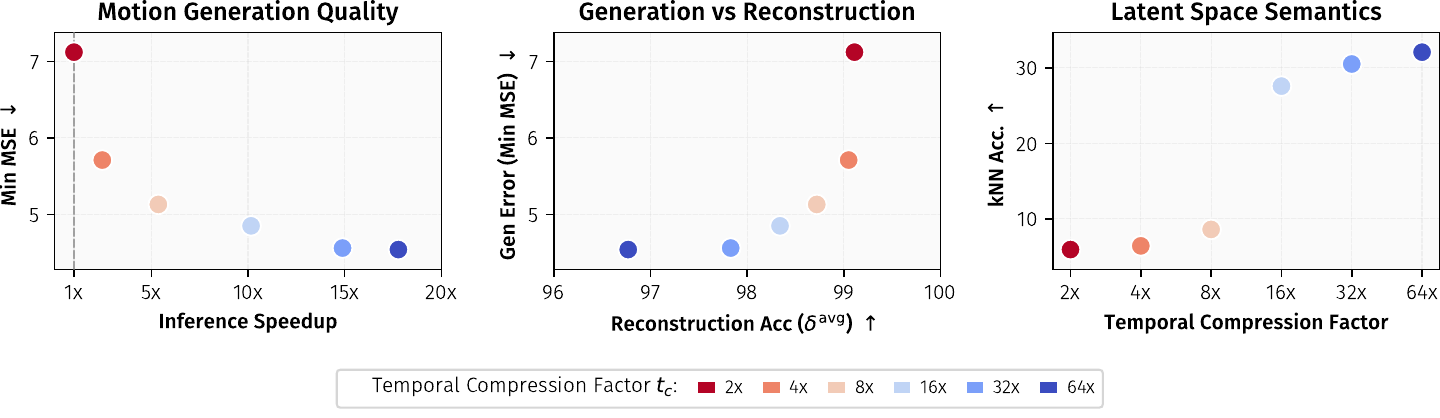}
    \caption{Temporal compression enables our model to generate plausible motions more efficiently. Under a fixed compute budget, both motion generation quality and inference throughput improve substantially with stronger compression (left), with only a minor reduction in reconstruction fidelity (middle). We attribute these gains to the reduced token count, which improves training efficiency, and to an increasingly semantic latent structure, as evidenced by higher kNN retrieval accuracy (right).    }
    \label{fig:vae_results}
\end{figure*}

\subsection{\todo{Reasoning in Motion Spaces}}
\label{method:planner}

We train a conditional flow matching model that operates directly in the latent motion space learned by the first stage. The goal is to model a distribution over motion latents $\mathbf{z}$ that captures the diversity of plausible trajectories while allowing for flexible conditioning on, e.g., text prompts or pokes~\cite{blattmann2021ipoke}. Given that the first stage already produces a compact and semantic motion embedding, learning to generate in this space becomes efficient.

Formally, we train a neural vector field $\mathbf{v}_\phi(\mathbf{z}_t, \mathbf{c}, t)$ to predict the instantaneous flow of a sample $\mathbf{z}_t$ along a continuous trajectory from a prior distribution $p_0(\mathbf{z}) = \mathcal{N}(0, I)$ to the empirical data distribution $p_1(\mathbf{z})$ of encoded motion latents. The conditioning variable $\mathbf{c}$ includes the start-frame embedding $\mathbf{f_0}$ and optionally additional control signals (language, pokes). The flow matching objective follows the continuous formulation of \cite{lipman2023flow}:
\begin{align}
\mathcal{L}_{\mathrm{FM}}(\phi) &=
\mathbb{E}_{t \sim \mathcal{U}(0,1)} \;
\mathbb{E}_{\mathbf{z}_0, \mathbf{z}_1 \sim p_0,\, p_1} \nonumber\\
&\quad [
\|
\mathbf{v}_\phi(\mathbf{z}_t, \mathbf{c}, t)
- \mathbf{v}_t^{\ast}(\mathbf{z}_0, \mathbf{z}_1)
\|_2^2
],
\end{align}
where $\mathbf{z}_t = (1-t)\mathbf{z}_0 + t\mathbf{z}_1$ is the linear interpolation between noise and data, and $\mathbf{v}_t^{\ast} = \mathbf{z}_1 - \mathbf{z}_0$ is the target flow field driving samples toward the data manifold. In practice, $\mathbf{v}_\phi$ is implemented as a transformer-based denoiser that takes as input the noisy latent $\mathbf{z}_t$, the scalar timestep $t$, and conditioning features $\mathbf{c}$. Both the poke and the text conditional model use cross attention to integrate the conditioning signal.
In the poke-conditional setting, we Fourier-embed the target poke positions and the target timestep, similar to~\cref{eq:enc}. The pokes' start positions are encoded using RoPE (\cref{eq:rope}). This design allows flexible conditioning on a variable number of pokes overall, as well as a variable number of pokes per trajectory, which can be placed arbitrarily throughout the temporal horizon.

%% file: sec/4_exps.tex
\section{Experiments}
We evaluate our model’s ability to compress motion and generate high-quality kinematics. First, we test our hypothesis that strong temporal compression enables efficient and accurate modeling of plausible motions. Second, we evaluate generative reasoning in this compressed motion space on both a closed-domain robotics setting and an open-domain general video setting. These experiments assess how well the model reasons about scene dynamics and plans trajectories that achieve goals specified through either text or poke.

\begin{figure}[t]
    \centering
    \includegraphics[width=0.32\linewidth]{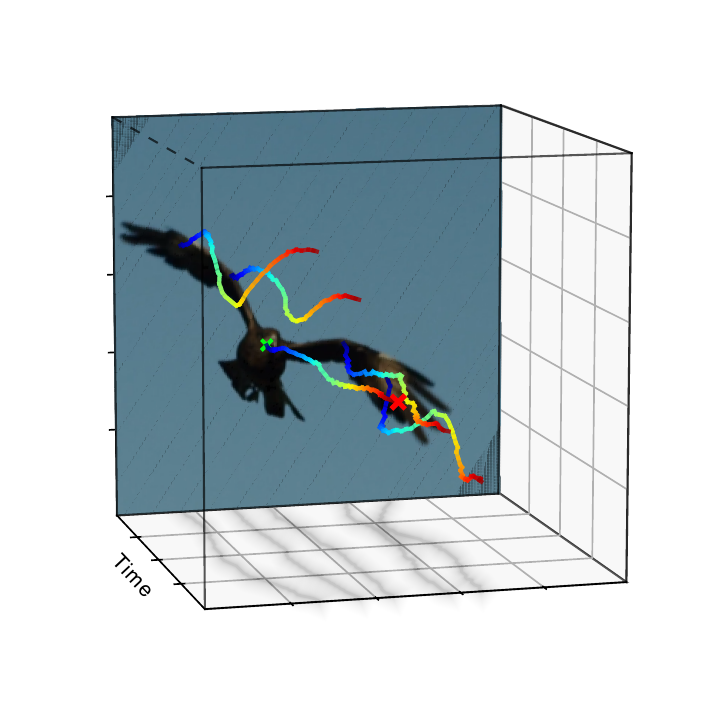}\hfill
    \includegraphics[width=0.32\linewidth]{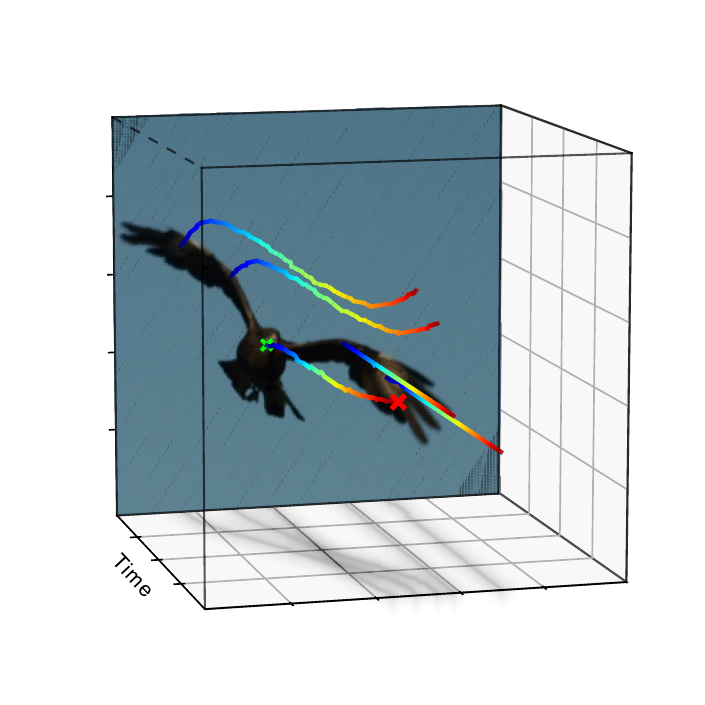}\hfill
    \includegraphics[width=0.32\linewidth]{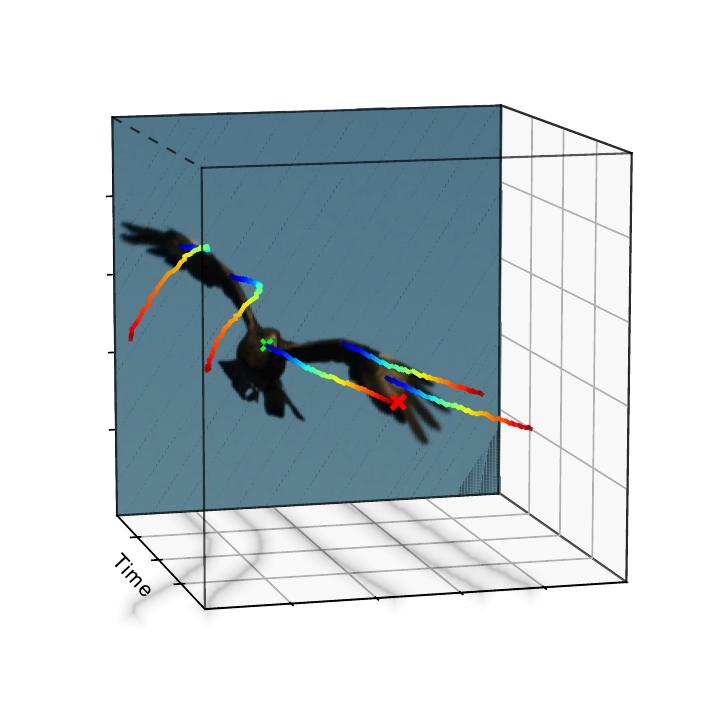}
    \caption{Example of multiple plausible motion hypotheses for the flight of an eagle, generated by our model from a single start frame. Our model produces diverse, physically coherent motions even in complex natural scenes, illustrating the expressiveness of the learned motion space.}
    \label{fig:egale}
\end{figure}

\begin{table*}[t]
    \centering

    \adjustbox{max width=\linewidth}{
    \begin{tabular}{lcc@{\hskip 1.5mm}c@{\hskip 1.5mm}c@{\hskip 1.5mm}cc@{\hskip 1.5mm}c@{\hskip 1.5mm}cc@{\hskip 1.5mm}c@{\hskip 1.5mm}cc@{\hskip 1.5mm}c@{\hskip 1.5mm}cc@{\hskip 1.5mm}c@{\hskip 1.5mm}c}
        \toprule
        \multirow{2}{*}[-3pt]{Method} & \multirow{2}{*}[-3pt]{Modality} & Speed & \multicolumn{3}{c}{1 Poke} & \multicolumn{3}{c}{2 Pokes} & \multicolumn{3}{c}{4 Pokes} & \multicolumn{3}{c}{8 Pokes} & \multicolumn{3}{c}{Dense} \\
         \cmidrule(lr){3-3} \cmidrule(lr){4-6} \cmidrule(lr){7-9} \cmidrule(lr){10-12} \cmidrule(lr){13-15} \cmidrule(lr){16-18}
        
        && timesteps$/$s $\uparrow$ & {\footnotesize Min $\downarrow$} & {\footnotesize Mean $\downarrow$} & {\footnotesize EPE $\downarrow$} & {\footnotesize Min $\downarrow$} &  {\footnotesize Mean $\downarrow$} & {\footnotesize EPE $\downarrow$}  & {\footnotesize Min $\downarrow$}  & {\footnotesize Mean$\downarrow$ } & {\footnotesize EPE $\downarrow$} & {\footnotesize Min $\downarrow$}  & {\footnotesize Mean $\downarrow$} & {\footnotesize EPE $\downarrow$} & {\footnotesize Min $\downarrow$}  & {\footnotesize Mean $\downarrow$} & {\footnotesize EPE $\downarrow$} \\
        \midrule
        
        {Motion-I2V~\cite{shi2024motioni2v} } & Flow & 21 & 135.7  & 429.7 & 19.7 & 106.5  & 336.2 & 17.1  & 87.4 & 245.7  & 14.4  & 54.5 & 121.7  & 11.2  & 46.9 & 71.7 &  8.8\\
        Track2Act~\cite{bharadhwaj2024track2act}& Tracks & 180 & - & - & - & - &- & - & -& - & - & - & - & - & 138.7 & 156.1 & 20.9 \\
        ZipMo (Ours) & Latent & \textbf{2500} 
        & \textbf{41.0} & \textbf{57.9} & \textbf{0.5} 
        & \textbf{40.9} & \textbf{57.2} & \textbf{0.5}  
        & \textbf{35.8} & \textbf{53.9} & \textbf{0.4}  
        & \textbf{34.8} & \textbf{48.7} & \textbf{0.7}  
        & \textbf{30.4} & \textbf{44.1} & \textbf{1.1} \\
        \bottomrule
    \end{tabular}
    }
    \caption{
    \textbf{Poked Motion Generation.} We compare against other methods that were trained on general video data and predict an explicit motion representation for multiple time steps. We report metrics for different conditioning densities to assess how well these models perform under varying levels of uncertainty. Track2Act~\cite{bharadhwaj2024track2act} is end frame conditional, which is why we only report numbers for the dense case. Our approach outperforms other models while also being significantly faster.
    }
    \vspace{-3mm}
    \label{tab:stage2}
\end{table*}

\paragraph{Implementation Details}
The model used in all experiments has a latent grid with spatial dimensions of $16\times16$ and a temporal compression of $64\times$. Both VAE and Motion Planner are implemented as transformers, generally following LLaMA~\cite{touvron2023llama}, with 340M and 530M parameters respectively. The goal conditioning is realized by cross-attending to either pokes for the open-domain setting or BERT~\cite{devlin2019bert} text embeddings, as is usual for LIBERO~\cite{wen2023any,yang2025tra}. For training, the models optimize their respective objective using AdamW~\cite{loshchilov2018decoupled} with $(0.9. 0.95)$ betas.
We provide additional implementation details of our model in \cref{sec:imp_details} and \cref{tab:imp_details}.

\paragraph{Datasets}
We consider both an open-domain and a closed-domain setting. For the open-domain setting, we train our models on KOALA-36M~\citep{wang2025koala}, aiming to capture broad world knowledge and diverse motion patterns. The source video clips are up to eight seconds long, and we downsample them by skipping every other frame, resulting in frame rates between 12 and 15 fps. Training supervision is provided by pseudo ground-truth trajectories obtained using TapNext ~\citep{zholus2025tapnext}. We filter out uncertain tracks as indicated by TapNext and train on tracks with 64 timesteps. We then evaluate on a subset of stock videos from Pexels~\citep{UmiMarch_OpenVideo_2025}, selected for scenes with a static camera that show complex motions and interactions, as introduced in~\cite{baumann2026envision}. See \cref{sec:tracker_abl} for an ablation on tracker choice.

In the closed-domain setting, we train and evaluate on CoTracker3~\cite{karaev2025cotracker3} tracks obtained from the LIBERO benchmark~\citep{liu2023libero}, which focuses on structured robotic interaction scenarios. Together, these datasets allow us to assess how well the learned motion spaces generalize across domains and motion regimes.

\paragraph{Metrics}
We evaluate generated motions in terms of plausibility, diversity, and distributional fidelity. Our goal is to measure how well a model captures physically and semantically plausible motions in a scene. Since real-world data typically provides only a single ground-truth motion per video, evaluation must capture distributional quality despite the lack of a ground-truth distribution. 
To that end, we adopt metrics from prior work, primarily WHN~\citep{boduljak2025happens}, focusing on measuring fidelity, plausibility, and \todo{diversity}.

To measure fidelity, we compute the minimum mean squared error (Min MSE), which measures the minimum distance between the ground truth (GT) motion and the distribution of generated samples. This metric correlates with the likelihood of the GT motion under the model’s distribution, as high-likelihood regions are sampled more densely, reducing the minimum MSE. Motion diversity is measured with the mean squared error to the GT averaged over all samples (Mean MSE). \todo{A Mean MSE close to the Min can be seen as evidence for a collapsed motion distribution whereas a very high Mean MSE indicates potential implausible motion and outliers.} Since our model is trained on open-set videos, the distribution of possible motions is highly multi-modal, with many possible motions being plausible under the model. Metrics assuming unimodal (ie, deterministic) correspondence are therefore unsuitable. In cases where we condition models on pokes, we additionally measure conditioning adherence with the endpoint error (EPE).
While no single metric independently guarantees meaningful motion generation, jointly considering Min MSE, Mean MSE, and EPE serves as a strong proxy for motion quality. All metrics are computed with tracks scaled to $[0, 128]$.

In the closed domain robotics setting, we measure the quality of generated motion embeddings by predicting a corresponding action sequence and executing it in simulation, allowing us to report success rates.

\subsection{Semantic Motion Compression }

We analyze the effectiveness of compressing motion across time using compression factors $t_c\in\{2,4,8,16,32,64\}$. A compression factor of $t_c$ means that $t_c$ consecutive frames are aggregated into a single latent representation with no temporal dimension. Temporal compression enables the model to represent long-term scene dynamics more efficiently while reducing computational cost.
We train models with varying temporal compression factors using a fixed compute budget. We evaluate the resulting motion generation quality in~\cref{fig:vae_results}. Results clearly demonstrate that high temporal compression consistently improves motion generation quality while also significantly increasing inference efficiency. 
Increasing temporal compression also enhances the semantic structure of the motion space. This is reflected in a monotonically increasing kNN accuracy on a subset of SSv2~\citep{goyal2017something}, shown in~\cref{fig:vae_results}. Higher kNN accuracies indicate that semantically similar motions are grouped more closely in the latent space.
We hypothesize that the improved motion generation quality under high temporal compression can be attributed to both the reduced token count, which increases training and inference efficiency, and a more semantically structured latent space.
Additional results comparing performance over the course of training are provided in~\cref{fig:compression_ablation_extended}, including two compute-matching schemes: matching either step time or batch size across models.

\subsection{Motion Reasoning and Planning}

We compare models in their ability to reason about physically plausible motions and plan trajectories that achieve desired goal states specified by prompts or end pokes. 

\subsubsection{Trajectory Prediction}

To comprehensively assess motion reasoning and planning, we conduct experiments in both a closed-domain robotics setting and an open-domain video setting.

\begin{figure}[t]
    \centering
    \begin{subfigure}{0.7\linewidth}
        \centering
        \includegraphics[width=\linewidth]{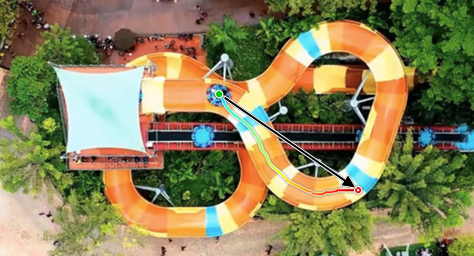}
        \caption{Path finding}
        \label{fig:pathfinding}
    \end{subfigure}
    
    \begin{subfigure}{0.7\linewidth}
        \centering
        \includegraphics[width=\linewidth]{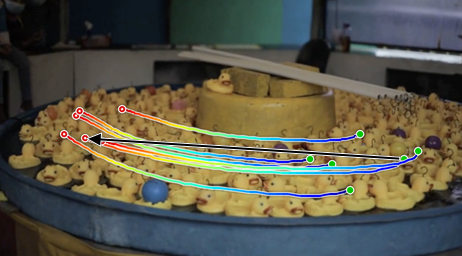}
        \caption{Rotational Motion}
        \label{fig:rotation}
    \end{subfigure}

    \begin{subfigure}{0.7\linewidth}
        \centering
        \includegraphics[width=\linewidth]{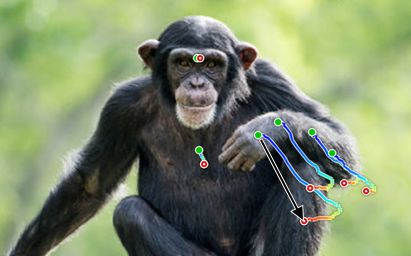}
        \caption{Joint Reasoning}
        \label{fig:rotation}
    \end{subfigure}

    \caption{Qualitative examples demonstrating diverse motion reasoning capabilities.
These results highlight that the model captures semantic motion structure and generalizes to varied motion regimes.
\vspace{-7mm}
}
    \label{fig:motion_examples}
\end{figure}

In the open-domain setting, we investigate whether models trained on large-scale, open-set data have acquired transferable knowledge about how objects move and interact in the world. This evaluation tests each model’s ability to generate physically plausible, goal-conditioned motions across diverse, unconstrained scenes.
We compare models that directly predict motion, either as trajectories or optical flow. Evaluation is performed in a conditional setup, where the model is given a start frame and either a set of $n$ target pokes or an end frame. The task is to generate plausible motions that transform the initial state into an end state consistent with the conditioning signals.
To assess how well models handle different levels of conditioning sparsity and therefore uncertainty, we vary the strength of the conditioning signal by using $n \in \{1, 2, 4, 8\}$ pokes, along with a dense setting that uses the strongest conditioning signal supported by each method. Sparse conditioning provides more flexibility but can lead to more implausible motions, whereas dense conditioning offers stronger guidance but increases the challenge of reconciling all conditionings into a consistent motion. As shown in~\cref{tab:stage2}, our model consistently outperforms both flow-based and trajectory-based baselines across all conditioning sparsities. Qualitative results in~\cref{fig:egale,fig:motion_examples} further demonstrate that our approach produces diverse, coherent, and physically plausible real-world motions. We show further results on DAVIS~\cite{davis} and PhysicsIQ~\cite{motamed2025generative} in~\cref{sec:additional_prediction}.

In the closed-domain setting, we evaluate our model's ability to generate trajectories from high-level text instructions on the LIBERO robotics dataset~\cite{liu2023libero}. Given a start frame and task description, the model predicts how the robot arm and surrounding objects should move to achieve the described goal, effectively performing motion planning from visual and linguistic input. Task descriptions in LIBERO are abstract~(see~\cref{fig:libero}), requiring the model to reason about both the current scene state and the specified goal to produce physically plausible and goal-directed trajectories. Results and evaluation details are presented in~\cref{app:traj_pred}.

\subsubsection{Action Prediction}

\begin{figure}
    \centering
    \begin{subfigure}{\linewidth}
        \caption{\textit{Turn on the stove and put the moka pot on it}}
        \includegraphics[width=0.23\linewidth]{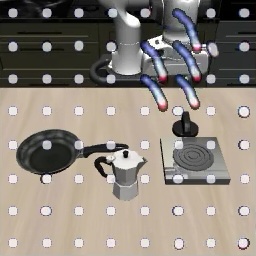}
        \hfill
        \includegraphics[width=0.23\linewidth]{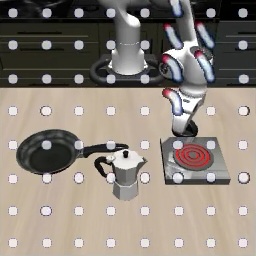}
        \hfill
        \includegraphics[width=0.23\linewidth]{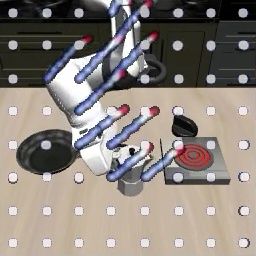}
        \hfill
        \includegraphics[width=0.23\linewidth]{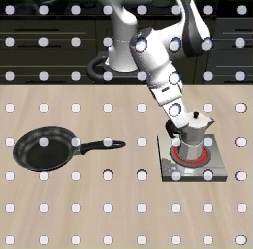}
    \end{subfigure}
    \begin{subfigure}{\linewidth}
        \caption{\textit{Put the yellow-and-white mug in the microwave and close it}}
        \includegraphics[width=0.23\linewidth]{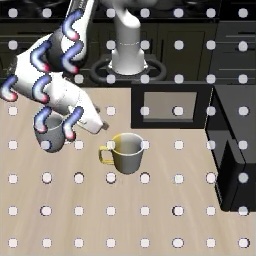}
        \hfill
        \includegraphics[width=0.23\linewidth]{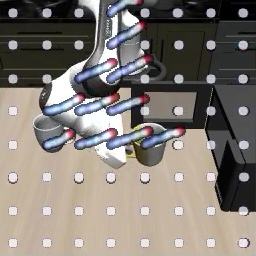}
        \hfill
        \includegraphics[width=0.23\linewidth]{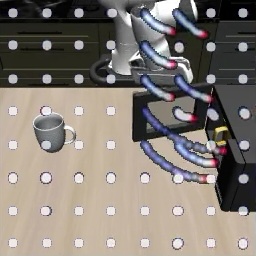}
        \hfill
        \includegraphics[width=0.23\linewidth]{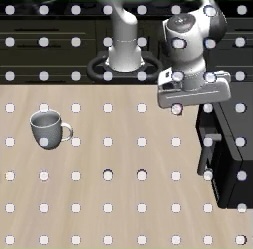}
    \end{subfigure}
    \caption{
    \textbf{LIBERO rollout samples.} Our track predictor forecasts tracks 16 steps ahead (visualized), enabling long-horizon planning. A policy head conditions on these predictions to select the next actions, with predictions updated after every new observation.
    \vspace{-3mm}
    }
    \label{fig:libero}
\end{figure}

\begin{table}[t]
    \centering
    \setlength{\tabcolsep}{3.6pt}
    \begin{tabular}{lcccccc}
        \toprule
        \multirow{2}{*}{Model} 
        & \multicolumn{6}{c}{LIBERO succ. rates $\uparrow$} \\ 
        \cmidrule(lr){2-7} 
         & 10 & 90 & Spatial & Goal & Object & Avg. \\ 
        \midrule

        ATM~\cite{wen2023any} & 39.3 & 48.4 & 68.5 & \textbf{77.8} & 68.0 & 60.4 \\
        Amplify~\cite{collins2025amplify} & 62.0 & \textbf{66.0} & 69.0 & 75.0 & 85.0 & 71.4 \\
        \(\text{ZipMo (Ours)}\) & \textbf{66.3} & 62.3 & \textbf{91.3} & 69.7 & \textbf{98.0} & \textbf{77.5} \\

        \midrule

        Tra-MoE~\cite{yang2025tra} & 28.5 & -- & 62.5 & 81.0 & 73.5 & 61.4 \\
        ZipMo (Ours) & \textbf{61.0} & -- & \textbf{85.7} & \textbf{82.3} & \textbf{92.0} & \textbf{80.3} \\

        \bottomrule
    \end{tabular}
    \caption{Success rates on LIBERO tasks following the respective training and evaluation setup of ATM~\citep{wen2023any} (top) and of Tra-MoE~\cite{yang2025tra} (bottom).}
    \label{tab:action_prediction_combined}
\end{table}

The LIBERO robotics setting allows us to rigorously evaluate generated motion embeddings by predicting corresponding robot actions, executing them in simulation and measuring task success. Following ATM~\citep{wen2023any} and Tra-MoE~\citep{yang2025tra}, we train a policy head that predicts robot actions from our generated conditional motion embeddings. 
Our model maintains a rolling motion plan over a horizon of $T$ steps, updated at each timestep with new observations. At timestep $t$, it predicts a motion embedding encoding future motion for $\tilde{t}\in[t, t+T]$ (see~\cref{fig:libero}), from which the policy head predicts the action at $t+1$. After executing this action, the model replans over the next horizon given the new observation.
Since the policy head receives only the generated motion embeddings as input, it learns a mapping from motion embeddings to actions, effectively acting as an \textit{inverse dynamics module}. This clear split ensures that the actual task-conditional reasoning and planning is done by the motion planner, not the policy head.
We follow the evaluation setup of both baselines and report results in~\cref{tab:action_prediction_combined}. Our model significantly outperforms both baselines across tasks, indicating superior understanding of scene dynamics. Further details are provided in~\cref{app:action_pred}.

\subsection{Comparison to Generative Video Models}
\label{sec:sota_eval}
To assess the effectiveness of modeling motion directly in our learned trajectory space, as opposed to modeling individual frames, we compare against state-of-the-art generative video models that also aim to learn world dynamics. The key distinction is that video models must jointly synthesize both appearance and motion, while our approach focuses exclusively on kinematics. Since we are interested in goal-directed motion rather than arbitrary dynamics, we focus on video models that can be conditioned on both a start frame and an end frame. This setup aligns with our conditioning scheme, where the model receives a start-frame embedding and a motion goal specified by pokes.

A direct comparison between our method and video models is challenging because videos do not provide explicit motion trajectories or any other motion representations. To obtain a comparable representation, we track the generated videos using CoTracker3~\cite{karaev2025cotracker3} (see \cref{sec:why_cotracker} for TapNext), producing estimated trajectories $\mathrm{tracks}_{\mathrm{gen}} \in \mathbb{R}^{N\times T\times 2}$.
However, tracking generated videos can introduce several sources of error, such as losing correspondences over time, resulting in incomplete or inconsistent motion estimates. In contrast, our method directly generates motion embeddings, providing globally consistent motion information without the risk of tracker drift or loss.

All methods are evaluated on a subset of Pexels~\cite{UmiMarch_OpenVideo_2025} videos, which contain diverse and visually distinct motions. While this subset was held-out for our models, we cannot guarantee that this was also the case for the video models. For each video, we extract ground-truth trajectories by initializing trackers on the start frame and tracking them across time until the end frame, retaining only moving tracks that remain visible throughout. This yields a ground-truth trajectory set $\mathrm{tracks}_{\mathrm{gt}} \in \mathbb{R}^{N\times T\times 2}$. Our model is conditioned on $N=40$ pokes $\mathbf{p}_{0\rightarrow T} \in \mathbb{R}^{N\times 2}$, each specifying the intended displacement from the start to the end frame, while video baselines are conditioned on the same start frame and the ground-truth end frame.

We report quantitative results using our proposed motion metrics. Because sampling from video diffusion models is computationally expensive, we consider two evaluation regimes:
(i) \textit{Sample Matched} where both methods generate the same number of samples, and (ii) \textit{Time Matched} where the wall-clock time for sampling is held constant, based on the average time required to generate a single video sample from the baseline. In the \textit{Sample Matched} regime, we outperform Wan, with a larger gap for Veo. We hypothesize that this is because Veo tends to generate very high magnitude motions, which yields high errors if the ground truth video has more moderate motion.

\begin{table}[t]
    \centering
    \rowcolors{2}{gray!10}{white}
    \resizebox{\linewidth}{!}{
    \begin{tabular}{lcccc}
        \toprule
        Model & Time $\downarrow$ &Min MSE $\downarrow$ & Mean MSE $\downarrow$ & EPE $\downarrow$ \\ %
        \midrule
        Wan~\cite{wan2025wan} & 1h &28.67& 57.02 & 4.68 \\ 
        Veo 3~\cite{Veo3} &?& 36.18 & 94.00 & 6.21\\
        \textbf{ZipMo (Ours)} & \textbf{1s} &\textbf{27.08} & \textbf{39.53} & \textbf{1.17} \\
     \bottomrule
    \end{tabular}
    }
    \caption{\textbf{Samples Matched:} We sample $k=8$ times from each model and track the generated videos to report our distributional metrics as well as conditioning adherence with EPE. This is an unfavorable setting for us, as our model is much smaller. Still, we outperform the video while being orders of magnitude faster.
    \vspace{-2mm}
    }
    \label{tab:sota-eval-sample}
\end{table}

\begin{table}[t]
    \centering
    \rowcolors{2}{gray!10}{white}
    \resizebox{\linewidth}{!}{
    \begin{tabular}{lcccc}
        \toprule
        Model & Samples $\uparrow$& Min MSE $\downarrow$ & Mean MSE $\downarrow$ & EPE $\downarrow$ \\ %
        \midrule
        Wan~\cite{wan2025wan} & 1 &64.20&64.20&5.23\\ 
        Veo 3~\cite{Veo3} & 1 & 65.99 & 65.99& 5.84\\
        \textbf{ZipMo (Ours)} & $\mathbf{>10k}$ & \textbf{21.29} & \textbf{40.33} & \textbf{1.17} \\
     \bottomrule
    \end{tabular}
    }
    \caption{\textbf{Time Matched:} Setup similar to \cref{tab:sota-eval-sample} but now matching wall clock time for sampling. Our performance lead increases drastically due to the efficiency of our approach. 
    \vspace{-5mm}
    }
    \label{tab:sota-eval-time}
\end{table}

%% file: sec/5_conclusion.tex
\section{Conclusion}

We have introduced a framework that enables highly efficient modeling of scene dynamics by operating on a learned motion embedding. By leveraging large-scale tracker-derived trajectories, our approach captures long-term kinematic structure while heavily compressing temporal information by $64\times$. Operating on these motion embeddings, our conditional flow matching model efficiently generates realistic, long-horizon motions that fulfill goals specified via text prompts or spatial pokes.
We empirically show that heavy temporal compression aids learning kinematics and enables our model to reason efficiently over long time. Across multiple benchmarks, our generated motion distributions significantly outperform those of both modern video models and task-specific approaches, demonstrating that our motion embeddings provide a compact, expressive, and flexible representation for modeling kinematics.

%% file: sec/X_suppl.tex
\clearpage
\setcounter{section}{0}
\renewcommand{\thesection}{\Alph{section}}

\renewcommand{\thefigure}{\Alph{figure}}
\renewcommand{\thetable}{\Alph{table}}

\maketitlesupplementary

\section{Additional Evaluation Details}

\paragraph{LIBERO Trajectory Prediction}
\label{app:traj_pred}

We compare the accuracy of our model's predicted motions against both discriminative and generative baselines in~\cref{tab:libero_eval}. Discriminative methods output a single set of trajectories for a given start frame and text prompt, while generative methods output distributions of possible trajectories. Across all tasks, our model significantly outperforms both discriminative and generative baselines, demonstrating superior understanding of the scene dynamics and the task at hand. For the generative methods, we report the Min MSE metric from WHN~\citep{boduljak2025happens} for $k=8$ samples (\textit{Min}) and for fair comparison with discriminative methods, also for $k=1$ (\textit{Single}). 

We closely follow the evaluation protocol of previous methods~\citep{wen2023any, yang2025tra, boduljak2025happens}: We split the videos into temporal windows of length $T$. Per temporal window, we extract $n=32$ trajectories that exceed a certain variance threshold, i.e., are non-static. Given the start frame of the temporal window and the $n$ starting positions, models have to predict the trajectories for the following $T$ frames. All existing methods use a temporal window size of $T=16$~\citep{wen2023any, yang2025tra, boduljak2025happens}. Since our model predicts 64 frames, we interpolate the baseline window of 16 frames up to 64 frames during training. At inference, we subsample our model’s prediction from 64 frames down to 16 frames to match the evaluation protocol. Pseudo-GT tracks are obtained through the exact preprocessing pipeline used by baselines~\cite{wen2023any, yang2025tra, boduljak2025happens}. The MSE metrics are computed on a resolution of $128 \times 128$. We train a single model jointly on both task suites (LIBERO-90 and LIBERO-10) and both views (side and effector).

\begin{table}[h]
    \centering
    \rowcolors{6}{gray!10}{white} %
    \setlength{\tabcolsep}{4pt} %
    \begin{tabular}{lcccc}
        \toprule
        \multirow{2}{*}{Model} 
        & \multicolumn{2}{c}{LIBERO-90} 
        & \multicolumn{2}{c}{LIBERO-10} \\ 
        \cmidrule(lr){2-3} \cmidrule(lr){4-5}
         & Side & Effector & Side & Effector \\ 
        \midrule
        \multicolumn{5}{l}{\textit{Discriminative}} \\[2pt]
        ATM~\cite{wen2023any} & 47.82 & 123.01 & 59.10 & 131.58 \\
        Tra-MoE~\cite{yang2025tra} & 39.77 & 116.47 & 50.37 & 131.75 \\
        \midrule

        \multicolumn{5}{l}{\textit{Generative}} \\[2pt]       
         \(\text{WHN}_{\text{Single}}\)~\cite{boduljak2025happens}* & 17.89 & 57.64 & 26.18 & 63.47 \\
        \(\text{ZipMo}_{\text{Single}}\) & \textbf{8.83} & \textbf{45.23} & \textbf{10.73} & \textbf{46.27} \\
        
        \(\text{WHN}_{\text{Min}}\)~\cite{boduljak2025happens}* & 10.99 & 32.01 & 13.86 & 35.93 \\
        \(\text{ZipMo}_{\text{Min}}\) & \textbf{5.96} & \textbf{27.78} & \textbf{7.43} & \textbf{25.80} \\

        \(\text{WHN}_{\text{Mean}}\)~\cite{boduljak2025happens}* & 18.32 & 60.47 & 26.71 & 66.35 \\
        \(\text{ZipMo}_{\text{Mean}}\) & \textbf{9.18} & \textbf{45.71} & \textbf{9.05} & \textbf{46.55} \\
        \bottomrule
    \end{tabular}
    \caption{Comparison on text-conditioned trajectory prediction on the LIBERO~\cite{liu2023libero} dataset, measured by MSE and following the evaluation setup of~\cite{boduljak2025happens}. Numbers marked with an asterisk~\textbf{*} are taken directly from the original paper, as no official checkpoints or training code were released. All other results were reproduced using the official checkpoints. The upper part of the table reports regression-based methods, while the lower part compares generative methods. The $_\text{Single}$ metric is the MSE for the first sample. All distributional metrics are computed over $k=8$ samples per trajectory.}
    \label{tab:libero_eval}
\end{table}

\paragraph{LIBERO Action Prediction}
\label{app:action_pred}

We train a policy head, also sometimes referred to as \textit{Inverse Dynamics Module}, to predict robot actions from our generated motion embeddings. In the LIBERO setting, actions are 7-dimensional vectors. We instantiate the policy head as a shallow 6-layer transformer with dimensionality 768. During policy head training and policy rollouts, we perform 10 sampling steps for the motion planner to generate the conditional motion embeddings. The policy head cross-attends only to these motion embeddings, thus we can be sure that the planning is actually done by the motion planner. Following ATM~\citep{wen2023any} and Tra-MoE~\citep{yang2025tra}, we train the policy head with a simple L2 regression loss, effectively performing Behavioural Cloning (BC). We train the policy head for 12 hours on 16 H200s. For both the ATM~\citep{wen2023any} and Tra-MoE~\citep{yang2025tra} baselines, we also follow their respective data configuration for training the motion planner. Unlike ATM~\citep{wen2023any}, we train a single joint motion planner across all LIBERO task suites, which further increases task diversity and planning complexity for our model. To increase robustness of the motion planner during policy rollouts, we lock the DINOv2 image encoder and apply noise augmentation to the start frame during training. Further, we condition the motion planner on the current episode timestep and randomly drop out the start frame as an additional incentive to follow the task conditioning. As LIBERO rollouts are non-deterministic due to both stochasticity in the simulation environment and the model predictions, we report mean and standard deviation across seeds in~\cref{tab:action_prediction_combined_with_ours_std}.

\begin{table*}[t]
    \centering
    \setlength{\tabcolsep}{8pt}
    \begin{tabular}{lcccccc}
        \toprule
        \multirow{2}{*}{Model} 
        & \multicolumn{6}{c}{LIBERO success rates $\uparrow$} \\ 
        \cmidrule(lr){2-7} 
         & 10 & 90 & Spatial & Goal & Object & Avg. \\ 
        \midrule

        ATM~\cite{wen2023any} & 39.3 & 48.4 & 68.5 & \textbf{77.8} & 68.0 & 60.4 \\
        Amplify~\cite{collins2025amplify} & 62.0 & \textbf{66.0} & 69.0 & 75.0 & 85.0 & 71.4 \\
        \(\text{ZipMo (Ours)}\) & \textbf{66.3 $\pm$ 4.0} & 62.3 $\pm$ 0.4 & \textbf{91.3 $\pm$ 0.6} & 69.7 $\pm$ 2.5 & \textbf{98.0 $\pm$ 2.6} & \textbf{77.5 $\pm$ 1.8} \\

        \midrule

        Tra-MoE~\cite{yang2025tra} & 28.5 & -- & 62.5 & 81.0 & 73.5 & 61.4 \\
        ZipMo (Ours) & \textbf{61.0 $\pm$ 1.7} & -- & \textbf{85.7 $\pm$ 2.1} & \textbf{82.3 $\pm$ 3.5} & \textbf{92.0 $\pm$ 2.0} & \textbf{80.3 $\pm$ 1.4} \\

        \bottomrule
    \end{tabular}
    \caption{Same setup as Table~\ref{tab:action_prediction_combined}, with seed-wise variability reported for our method. For each experiment, we evaluate three random seeds, with 10 rollouts per task per seed, and report mean success rate $\pm$ standard deviation.}
    \label{tab:action_prediction_combined_with_ours_std}
\end{table*}

\paragraph{Video Models}
We compare our approach against two state-of-the-art generative video models that offer start and end frame conditioning, making them suitable baselines for our goal-conditional motion generation task. For Wan~\cite{wan2025wan}, we use the \texttt{Wan-AI/Wan2.1-FLF2V-14B-720P-Diffusers} implementation from Hugging Face and run it locally. Since Veo 3 is a closed-source model, we obtain samples via the fal.ai API. Due to the high computational and monetary cost associated with sampling from these models, we focus our evaluation on a curated subset of the Pexels dataset~\cite{UmiMarch_OpenVideo_2025}, selecting 68 videos that feature unconstrained, real-world motion diversity.

To ensure a meaningful evaluation of motion generation, we carefully selected videos in which the primary moving objects remain clearly visible throughout the entire sequence. Our curated set features a diverse range of scenes, spanning various object categories including humans engaged in different activities, animals, vehicles, natural landscapes, and urban environments. The number of moving objects varies across videos, encompassing both single-object and multi-object scenarios. We further ensured diversity in motion by including clips with a wide spectrum of motion magnitudes, ranging from subtle, fine-grained movements to pronounced, large-scale motions.

Each original Pexels video has a framerate between 24 and 30 fps. We subsample each video with a frame skip of 1, selecting a contiguous window of 64 frames, corresponding to roughly 4-5 seconds of video. To establish ground-truth motion, we randomly sample 1024 query points in the first frame of each video and track them forward throughout the sequence. From these, we randomly select 40 dynamic (i.e., non-static) tracks, which serve as our final ground-truth trajectories for evaluation.

For both video models, we condition generation on the first and last frame of each clip, aligning the setup with our poke-conditional evaluation. The Wan model produces 81 frames\footnote{81 frames is the minimum number of frames that this version of Wan can generate. Shorter videos are not possible.} at 12 fps, while Veo 3 generates 96 frames at 24 fps. To account for stochasticity and fairly assess distributional metrics, we sample $K=8$ generations per start–end frame pair, resulting in a total of $68 \cdot 8 = 544$ generated video samples per model.

We then track the same 40 query points in each generated video using the same tracking procedure as was used to establish the ground truth tracks. Since both models output more frames than our ground truth sequences, we downsample the resulting tracks to match the 64-frame ground truth length. Finally, we compute the evaluation metrics outlined in the experiments section, comparing the predicted tracks from the generated videos to the ground truth to assess the quality and diversity of motion generation.

\section{Implementation Details}
\label{sec:imp_details}
We train our model in two stages: a variational autoencoder (VAE) that compresses tracking data into a latent grid representation, and a flow matching model that generates these latent representations conditioned on task specifications. Table~\cref{tab:imp_details} summarizes the key hyperparameters for both stages. Both models are trained on 10M video clips from diverse open-set videos, using TAPNext~\cite{zholus2025tapnext} to extract 1024 randomly sampled tracking positions per clip, with poke coordinates normalized to the range [-1,1]. 

The VAE employs a dual-encoder architecture with 12 layers each for self-attention and cross-attention processing, where tracking tokens attend to image tokens via cross-attention. Image features are extracted using a pretrained DINOv2 ViT-B/14 encoder~\cite{oquab2023dinov2}, which remains frozen during initial training and is unlocked later when scaling to larger batch sizes. Tracker points are first Fourier embedded and processed with an MLP before passing them to the transformer. The VAE uses 3D RoPE~\cite{su2024roformer,crowson2024hdit} as positional encoding, compressing the tracking data into a 16$\times$16 latent grid with 16 channels. We set the KL divergence weight to $\beta = 1.0 \times 10^{-7}$ to regularize the latent space and use an L1 reconstruction loss. During early experiments, we evaluated alternative reconstruction losses (L2 and Huber) and alternative prediction targets (auto-regressive deltas and offsets to the start location). Directly predicting absolute coordinates in our normalized space with an L1 loss consistently gave the best performance. The VAE is trained with AdamW~\cite{loshchilov2018decoupled} using a constant learning rate of $1.0\times 10^{-4}$ and stable decay (WSD) \cite{hu2024minicpm}, gradually scaling the batch size from 64 to 256 over 800k steps. 

The second stage motion planner is a 24-layer transformer with 1024-dimensional self-attention and cross-attention, trained to perform flow matching in the learned latent space. Unlike the VAE, this model processes both image tokens and latent grid tokens through shared self-attention layers, while task specifications (either Fourier-embedded pokes or text embeddings) are provided via cross-attention. The motion planner uses 2D RoPE for positional encoding, as it operates on the 2D latent grid output by the VAE. For ablation experiments, we also evaluate using 3D RoPE in the motion planner in case the latent grid has a temporal dimension. Training follows a similar schedule to the VAE, scaling from 512 to 2048 batch size over 700k steps with a constant learning rate. Both models use RMSNorm\cite{zhang2019root}, SwiGLU activations~\cite{shazeer2020glu}, and an FFN expansion factor of 3, and are trained in bfloat16 mixed precision on 16 to 64 Nvidia H200 GPUs, each requiring approximately 3 days of training time. 

\begin{figure}[t]
    \centering
    \begin{subfigure}{0.7\linewidth}
        \centering
        \includegraphics[width=\linewidth]{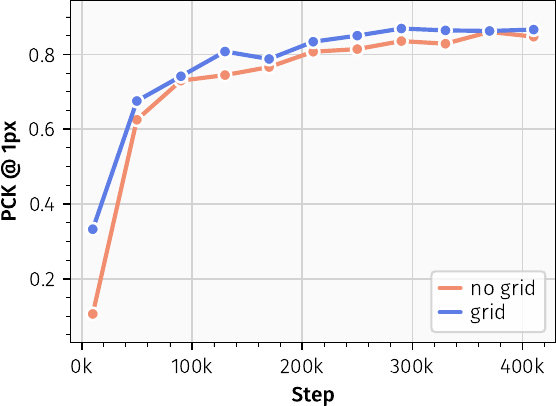}
        \caption{AE PCK}
    \end{subfigure}\\[1em]
    \begin{subfigure}{0.7\linewidth}
        \centering
        \includegraphics[width=\linewidth]{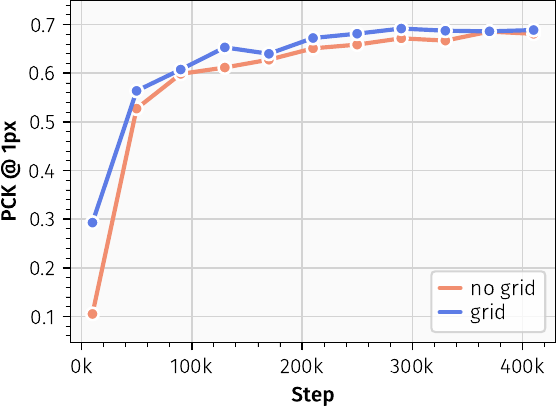}
        \caption{MAE PCK}
    \end{subfigure}
    \caption{We ablate the effect of arranging latent tokens in a grid using 2D RoPE. We measure PCK using a 1 pixel threshold and report reconstruction performance for encoded positions (AE setting), as well as unseen positions (MAE setting). We find that providing positional information slightly increases performance. }
    \label{fig:grid_ablations}
\end{figure}

\begin{figure}[t]
\centering
        \includegraphics[width=.7\linewidth]{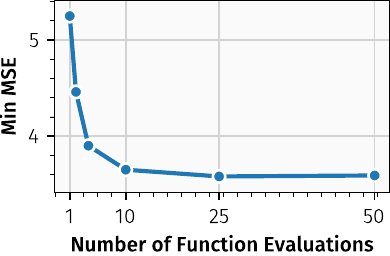}
        \caption{Motion Generation performance (Min MSE) across different numbers of function evaluations (NFEs). Trajectories are densely conditioned on target positions, sampled $k=128$ times per start frame.}

    \label{fig:nfe_ablation}
\end{figure}

We conduct two ablations on the model design. ~\cref{fig:grid_ablations} demonstrates that explicitly arranging latent tokens in a grid structure leads to slightly better performance compared to treating them as an unstructured sequence. ~\cref{fig:nfe_ablation} shows the performance of the flow matching model across different numbers of function evaluations (NFEs), demonstrating that the model achieves decent performance even with as few as 10 sampling steps, making it practical for more time-sensitive applications. All ablation models use identical hyperparameters to the main models but are trained on only 4 Nvidia H200 GPUs for 24 hours to enable rapid experimentation.

\begin{table}[t]
    \centering
    \adjustbox{max width=\columnwidth}{
    \begin{tabular}{l@{}cc}
        \toprule
        Parameter & ZipMo VAE & ZipMo Planner \\
        \midrule
        Dataset & Open-Set Videos & Open-Set Videos \\
        Number of clips & 10M & 10M \\
        Tracker & TAPNext~\cite{zholus2025tapnext} & TAPNext~\cite{zholus2025tapnext} \\
        Tracker positions & 1024 random & 1024 random \\
        Poke scale & $[-1, 1]$ & $[-1, 1]$ \\
        \midrule
        Batch size & 64 $\rightarrow$ 256  & 512 $\rightarrow$ 2048  \\
        Optimizer & AdamW~\cite{loshchilov2018decoupled} & AdamW~\cite{loshchilov2018decoupled} \\
        Peak LR & $1.0 \times 10^{-4}$ & $1.0 \times 10^{-4}$ \\
        LR schedule & constant with WSD~\cite{hu2024minicpm} & constant \\
        Betas & $(0.9, 0.95)$ & $(0.9, 0.95)$ \\
        Warm-up steps & 300 & 300 \\
        Total Steps & 800k & 700k \\
        Precision & bfloat16 & bfloat16 \\
        Total Parameters & 340M & 530M \\
        GPUs & 16 $\rightarrow$ 64 Nvidia H200 & 16 $\rightarrow$ 64 Nvidia H200 \\
        Training Time & 3 days & 3 days \\
        \midrule
        Depth & 12 \& 12 & 24 \\
        SA width & 768 & 1024 \\
        CA width & 768 & 1024 \\
        Normalization & RMSNorm~\cite{zhang2019root} & RMSNorm~\cite{zhang2019root} \\
        FFN expand factor & 3 & 3\\
        Activation & SwiGLU~\cite{shazeer2020glu} & SwiGLU~\cite{shazeer2020glu} \\
        Positional encoding & 3D RoPE ~\cite{su2024roformer,crowson2024hdit} & 2D RoPE ~\cite{su2024roformer,crowson2024hdit} \\
        Image size & $224 \times 224$ & $224 \times 224$ \\
        Image encoder & DINOv2 ViT-B/14~\cite{oquab2023dinov2} & DINOv2 ViT-B/14~\cite{oquab2023dinov2} \\
        Latent width & 16& 16 \\
        Latent size &16$\times$16&16$\times$16 \\
        KL loss $\beta$ &$1.0 \times 10^{-7}$& -\\
        \bottomrule
    \end{tabular}
    }
    \caption{Hyperparameters for our first stage VAE and motion planner. Ablation models use the same parameters, but were only trained on 4 Nvidia H200 GPUs for 24 hours.}
    \label{tab:imp_details}
    \vspace{-5mm}
\end{table}

\section{Track Densification and Inpainting}
\begin{figure}[t]
    \centering
    \begin{subfigure}{\linewidth}
        \centering
        \includegraphics[width=0.9\linewidth]{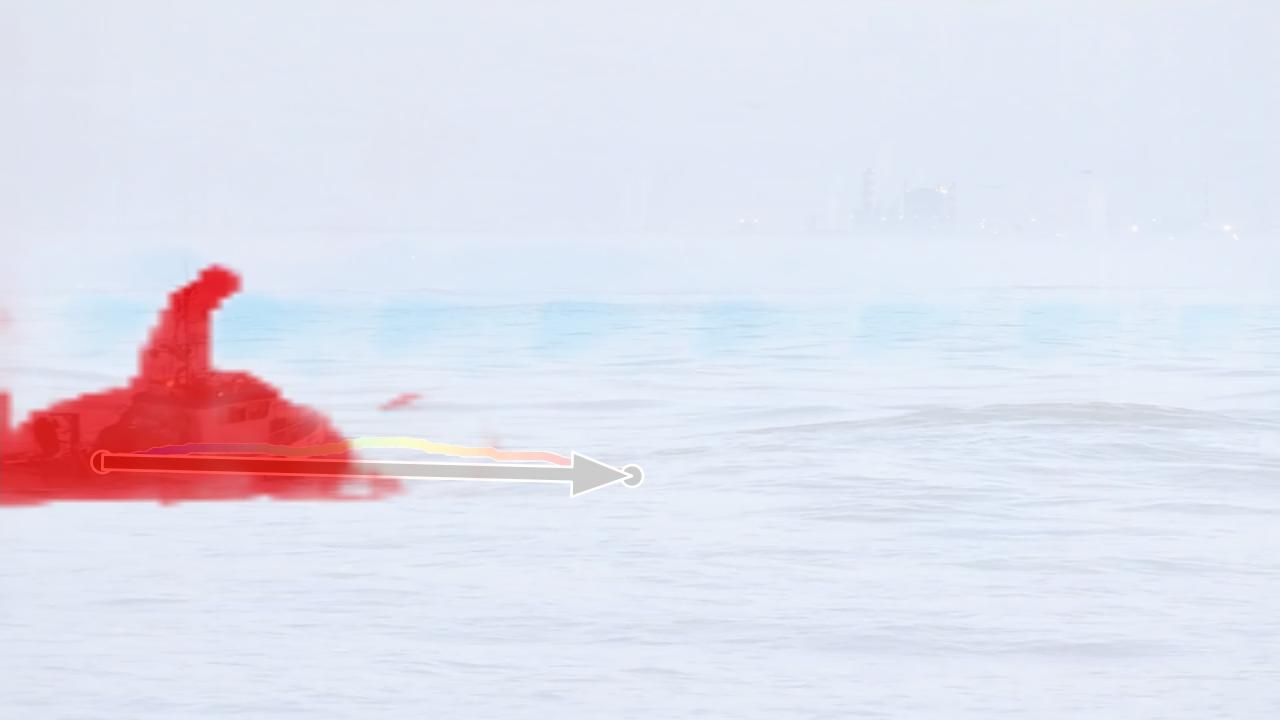}
    \end{subfigure}\\[1em]
    \begin{subfigure}{\linewidth}
        \centering
        \includegraphics[width=0.9\linewidth]{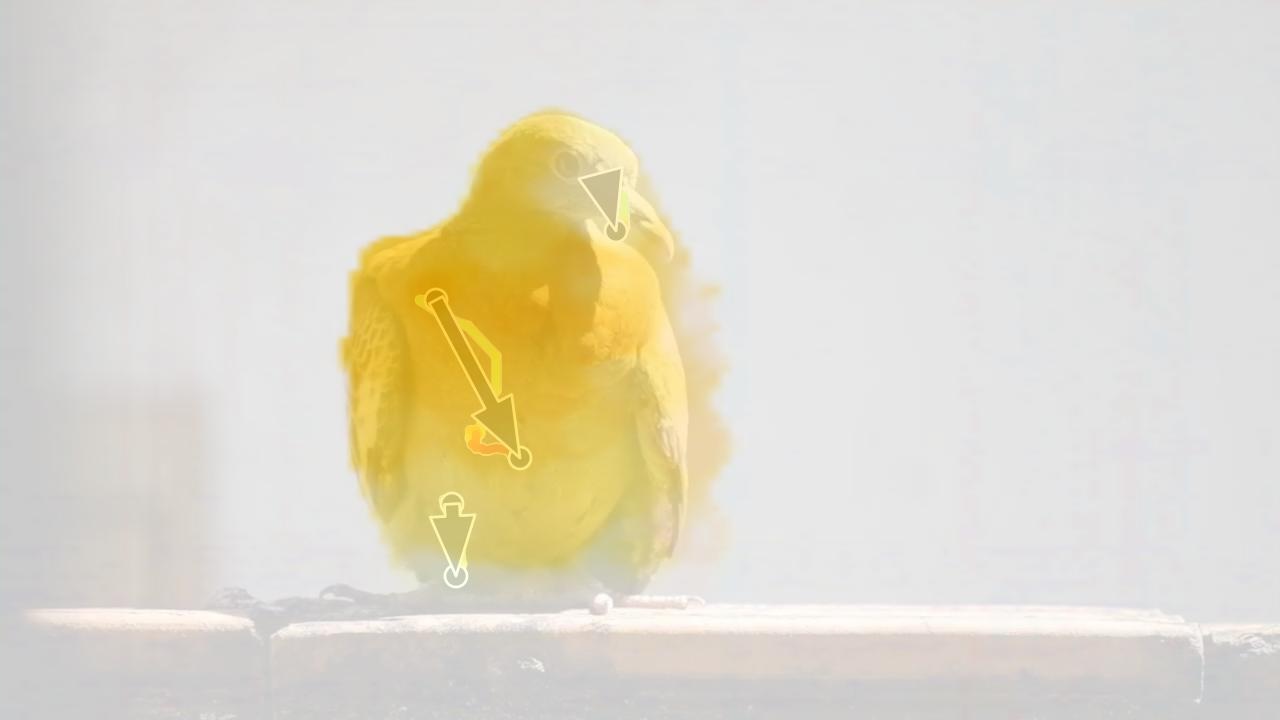}
    \end{subfigure}
    \caption{Our model enables dense motion inference from sparse tracker inputs by converting observed tracks (pokes) into conditioning signals that specify desired endpoints in the final frame. The visualizations show dense flow toward the goal frame, illustrating how our approach infers globally coherent motion. Only the endpoint for each poke is provided; the resulting dense prediction demonstrates our model’s ability to interpolate and inpaint plausible trajectories across the entire scene.}
    \label{fig:dense}
\end{figure}

Trajectories provide a sparse sampling of motion and do not capture every pixel in a video. However, some application require or benefit from a dense a dense motion representation. Our approach naturally supports track densification, as illustrated in \cref{fig:dense}.

This capability arises from two key properties of our framework. First, the motion planner (\cref{method:planner}) always generates a dense latent motion grid, regardless of how dense or sparse the poke conditioning is, due to our training setup. Second, our first-stage autoencoder (\cref{subsec:learning_motion_spaces}) can be queried at arbitrary spatial locations. Thus, by converting available tracks into pokes and conditioning the motion planner on these inputs, we can generate a dense latent grid and subsequently obtain densely inpainted tracks by querying the autoencoder at all pixel locations. This procedure enables our model to reconstruct dense motion fields from sparse tracker inputs, providing high-resolution inpainting of trajectories where needed.

\section{Effects of Temporal Compression}

\begin{figure}[t]
    \centering
    \begin{subfigure}{0.49\linewidth}
        \centering
        \includegraphics[width=\linewidth]{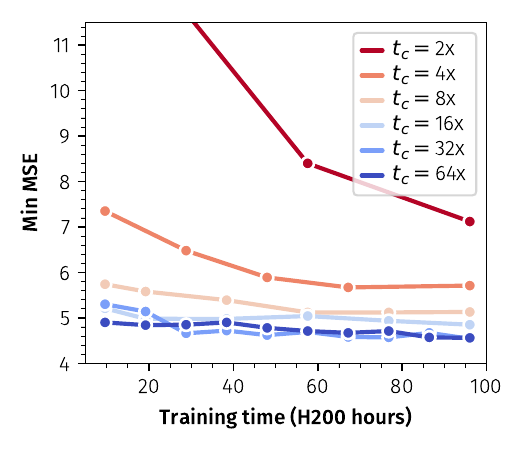}
        \caption{matched step time}
        \label{fig:r-vs-g}
    \end{subfigure}
    \hfill
    \begin{subfigure}{0.49\linewidth}
        \centering
        \includegraphics[width=\linewidth]{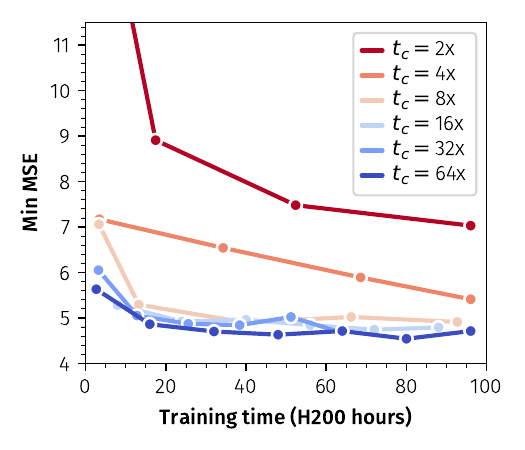}
        \caption{matched batch size}
        \label{fig:second}
    \end{subfigure}
    \caption{Motion generation quality over training for different temporal compression factors, measured against wall-clock time to account for varying computational cost. Matching step time (left) and batch size (right) both show that higher temporal compression yields more efficient learning and faster emergence of realistic motion.}
    \label{fig:compression_ablation_extended}
\end{figure}

We extend our analysis of the temporal compression factor's influence on generation performance from~\cref{fig:vae_results}. To that end, we show motion generation performance over the course of training in two compute-matched settings in~\cref{fig:compression_ablation_extended}: matching step time across models and matching batch size across models. As discussed in the main paper, the number of tokens to be denoised is inversely proportional to the temporal compression factor. To illustrate, the model with $t_c=2$ denoises $32\times$ as many tokens as the model with $t_c=64$, which drastically increases the compute per sample both during training and inference. Results clearly demonstrate that our strong temporal compression is beneficial for motion generation performance, while also being much more efficient at inference time (see~\cref{fig:vae_results}).

\section{Tracker Model Ablation}
\label{sec:tracker_abl}
To directly assess whether the learned motion space inherits tracker-specific biases, we ablate both the supervision source and its quality. Since the first stage embeds trackers into a smooth space, we report averaged reconstruction PCKs~($\delta^\mathrm{avg}$).
Training with TapNext versus CoTracker3 yields near-identical reconstruction accuracy, and models generalize well when evaluated with the other tracker (\cref{tab:cross_tracker}), indicating limited sensitivity to tracker choice.

We also stress-test supervision quality by degrading tracks during training.
Dropping out tracks or training only on non-occluded tracks leads to graceful degradation on an evaluation split that includes occlusions (\cref{tab:tracker_degradation}), indicating robustness to imperfect supervision. 
While we acknowledge that tracker quality theoretically upper-bounds performance, our method outperforms baselines on downstream tasks~(\cref{tab:action_prediction_combined}) and exceeds motion prediction performance of models trained with flow~(\cref{tab:stage2}) or RGB supervision~(\cref{tab:sota-eval-sample,tab:sota-eval-time}).

\begin{table}
    \begin{centering}
    \begin{tabular}{lcc}
        \toprule
        \textbf{Train $\backslash$ Eval} & \textbf{TapNext} & \textbf{CoTracker3} \\
        \midrule
        TapNext    & \cellcolor{ourblue!10}96.8 & \cellcolor{ourblue!20}97.0 \\
        CoTracker3 & 96.3 & \cellcolor{ourblue!40}97.3 \\
        \bottomrule
    \end{tabular}
    \captionof{table}{Cross-tracker reconstruction accuracy $\delta^\mathrm{avg}$.}
    \label{tab:cross_tracker}
    \end{centering}
\end{table}

\begin{table}
    \begin{centering}
    \begin{tabular}{lc}
                \toprule
                \textbf{Ablation} & $\delta^\textrm{avg}\uparrow$ \\
                \midrule
                Ours & 96.8 \\
                $+$ track dropout & 94.0 \\
                $+$ filter occlusions & 93.2 \\
                \bottomrule
            \end{tabular}
            \captionof{table}{Reconstruction accuracy degradation under systematic tracker degradation during training.}
            \label{tab:tracker_degradation}
    \end{centering}
\end{table}

\begin{table}[t]
    \centering
    \resizebox{\linewidth}{!}{
    \begin{tabular}{lccc}
            \toprule
            \textbf{Model} & \textbf{Min MSE} $\downarrow$ & \textbf{Mean MSE} $\downarrow$ & \textbf{EPE} $\downarrow$ \\
            \midrule
             \multicolumn{3}{l}{\textit{DAVIS~\cite{davis}}}\\
             Motion I2V~\cite{shi2024motioni2v} & 222.2  & 307.0 & 16.37 \\
             ZipMo (Ours) & \textbf{155.1}  & \textbf{233.0}  & \textbf{0.83}\\
             \midrule
             \multicolumn{3}{l}{\textit{PhysicsIQ~\cite{motamed2025generative}}}\\
             Motion I2V~\cite{shi2024motioni2v} & 177.8 & 225.1 &  12.4\\
             ZipMo (Ours) &  \textbf{90.60} &  \textbf{143.65} &  \textbf{0.76}\\
             \bottomrule
        \end{tabular}
    }
    \caption{
    Dense Track prediction results on Davis \textit{(top)} and PhysicsIQ \textit{(bottom)}. We condition both models on the maximum number of pokes from the start to the end frames. 
    }
    \label{tab:davis}
\end{table}

\section{Why CoTracker for Video Evaluation}
\label{sec:why_cotracker}
We use CoTracker3 to obtain tracks from the generated videos in our SOTA evaluation (\cref{sec:sota_eval}) because TapNext can lose tracks, resulting in missing values. This makes an unbiased challenging as it is unclear how to deal with these missing tracks. For completeness, we rerun the evaluation with tracks obtain from TapNext where we inpainting missing track values with the last observed position. This worsens metrics for video models (Min MSE$\downarrow$: Wan 29$\rightarrow$37, Veo~3 36$\rightarrow$48).

\section{Additional Track Prediction Results}
\label{sec:additional_prediction}
We focus on static scenes to avoid conflating detailed object motion \todo{since heavy camera motion would dominate the metrics}.%
We further evaluate our model on the DAVIS 2017 dataset, comprising 150 videos with often significant camera motion, as well as the solid mechanics split of Physics IQ, focusing on physical understanding.
In both settings, we outperform Motion I2V, our strongest direct competitor, in its strongest Dense setting (\cref{tab:davis}).

\section{Addional Qualitative Examples}
\begin{figure*}[]
    \centering

    \begin{subfigure}[b]{0.3\textwidth}
        \centering
        \includegraphics[width=\linewidth]{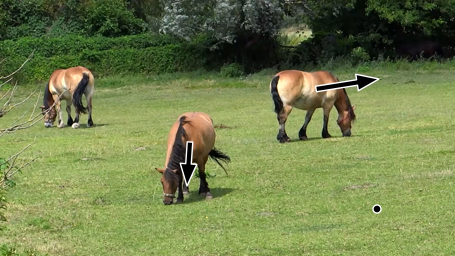}
    \end{subfigure}\hfill
    \begin{subfigure}[b]{0.3\textwidth}
        \centering
        \includegraphics[width=\linewidth]{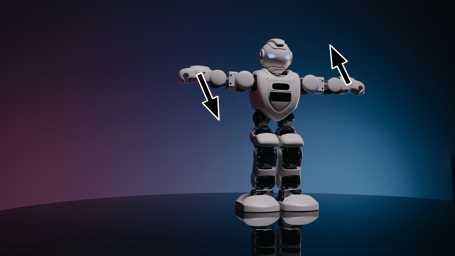}
    \end{subfigure}\hfill
    \begin{subfigure}[b]{0.3\textwidth}
        \centering
        \includegraphics[width=\linewidth]{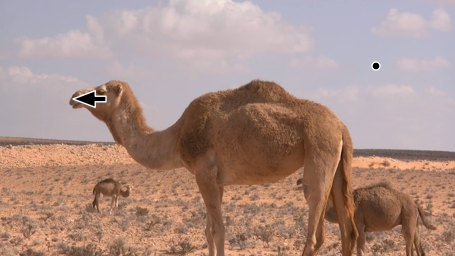}
    \end{subfigure}
    
    \begin{subfigure}[b]{0.3\textwidth}
        \centering
        \includegraphics[width=\linewidth]{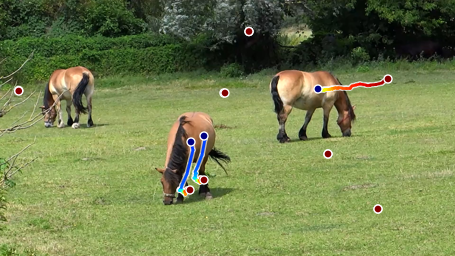}
    \end{subfigure}\hfill
    \begin{subfigure}[b]{0.3\textwidth}
        \centering
        \includegraphics[width=\linewidth]{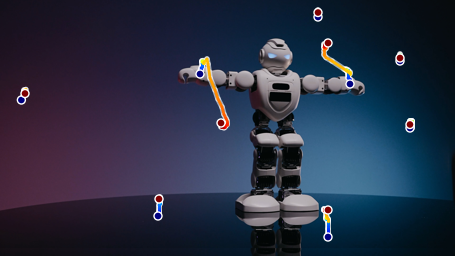}
    \end{subfigure}\hfill
    \begin{subfigure}[b]{0.3\textwidth}
        \centering
        \includegraphics[width=\linewidth]{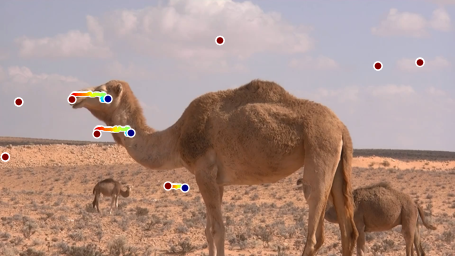}
    \end{subfigure}

    \begin{subfigure}[b]{0.3\textwidth}
        \centering
        \includegraphics[width=\linewidth]{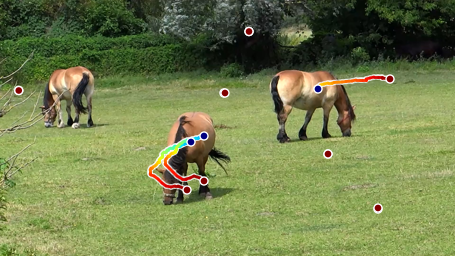}
    \end{subfigure}\hfill
    \begin{subfigure}[b]{0.3\textwidth}
        \centering
        \includegraphics[width=\linewidth]{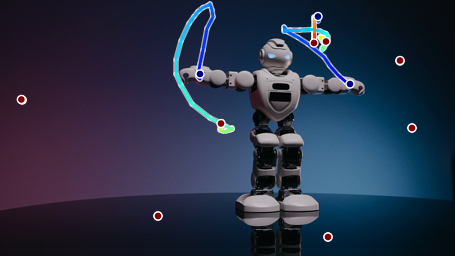}
    \end{subfigure}\hfill
    \begin{subfigure}[b]{0.3\textwidth}
        \centering
        \includegraphics[width=\linewidth]{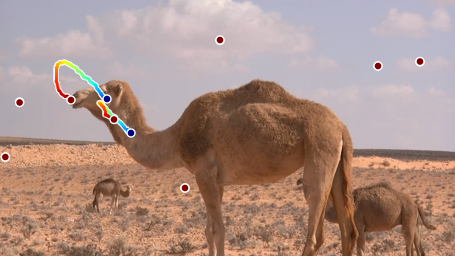}
    \end{subfigure}

    \begin{subfigure}[b]{0.3\textwidth}
        \centering
        \includegraphics[width=\linewidth]{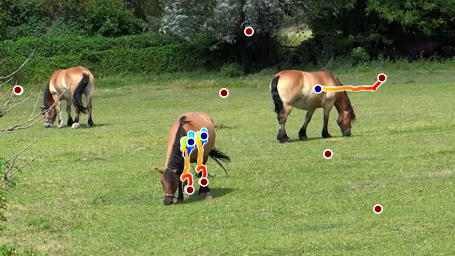}
    \end{subfigure}\hfill
    \begin{subfigure}[b]{0.3\textwidth}
        \centering
        \includegraphics[width=\linewidth]{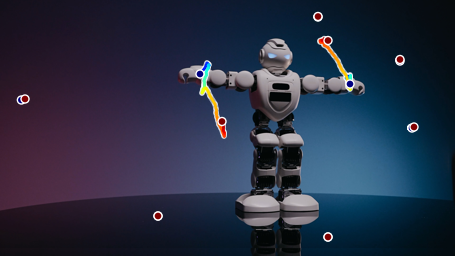}
    \end{subfigure}\hfill
    \begin{subfigure}[b]{0.3\textwidth}
        \centering
        \includegraphics[width=\linewidth]{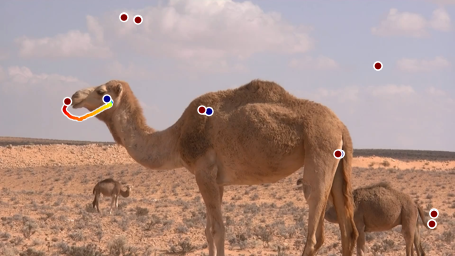}
    \end{subfigure}

    \begin{subfigure}[b]{0.3\textwidth}
        \centering
        \includegraphics[width=\linewidth]{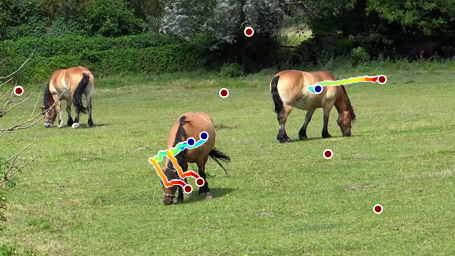}
    \end{subfigure}\hfill
    \begin{subfigure}[b]{0.3\textwidth}
        \centering
        \includegraphics[width=\linewidth]{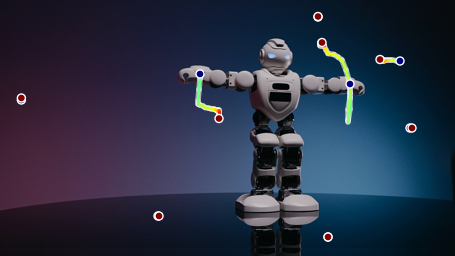}
    \end{subfigure}\hfill
    \begin{subfigure}[b]{0.3\textwidth}
        \centering
        \includegraphics[width=\linewidth]{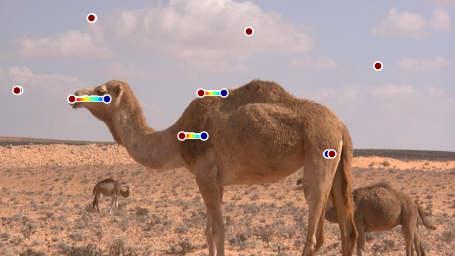}
    \end{subfigure}

    \caption{\textbf{Additional qualitative samples.} The first image in every column is the prompt, with arrows indicating pokes, i.e., where a specific start position should end. The remaining four images are samples from our models, illustrating trajectories of randomly selected query points. The color gradient of the trajectories acts as an indicator of how fast the motion is happening. The gradient starts at blue, moves to green, yellow, and finally ends at red.}
    \label{fig:app_qual_1}
\end{figure*}

\begin{figure*}[]
    \centering

    \begin{subfigure}[b]{0.3\textwidth}
        \centering
        \includegraphics[width=\linewidth]{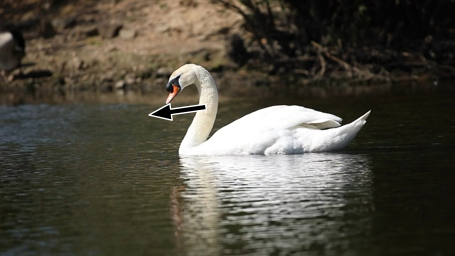}
    \end{subfigure}\hfill
    \begin{subfigure}[b]{0.3\textwidth}
        \centering
        \includegraphics[width=\linewidth]{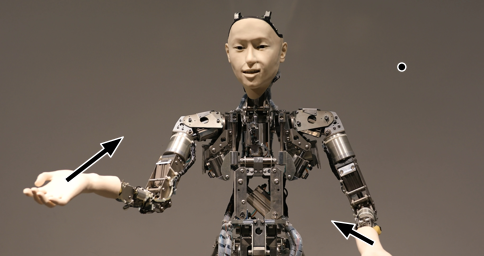}
    \end{subfigure}\hfill
    \begin{subfigure}[b]{0.3\textwidth}
        \centering
        \includegraphics[width=\linewidth]{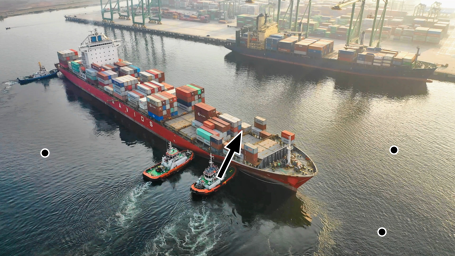}
    \end{subfigure}
    
    \begin{subfigure}[b]{0.3\textwidth}
        \centering
        \includegraphics[width=\linewidth]{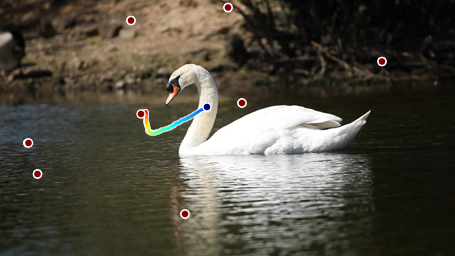}
    \end{subfigure}\hfill
    \begin{subfigure}[b]{0.3\textwidth}
        \centering
        \includegraphics[width=\linewidth]{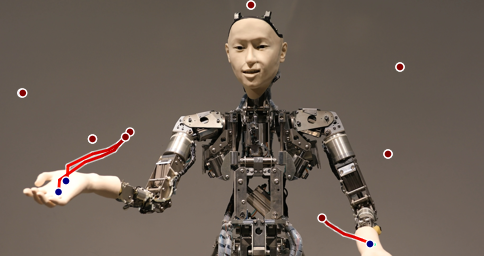}
    \end{subfigure}\hfill
    \begin{subfigure}[b]{0.3\textwidth}
        \centering
        \includegraphics[width=\linewidth]{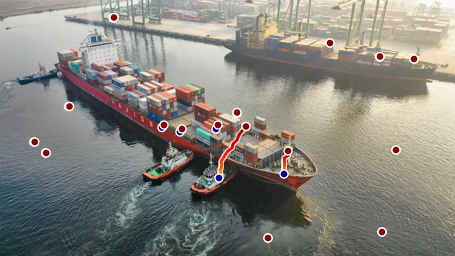}
    \end{subfigure}

    \begin{subfigure}[b]{0.3\textwidth}
        \centering
        \includegraphics[width=\linewidth]{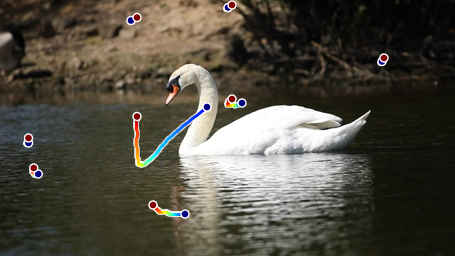}
    \end{subfigure}\hfill
    \begin{subfigure}[b]{0.3\textwidth}
        \centering
        \includegraphics[width=\linewidth]{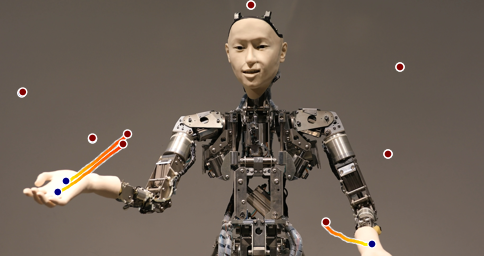}
    \end{subfigure}\hfill
    \begin{subfigure}[b]{0.3\textwidth}
        \centering
        \includegraphics[width=\linewidth]{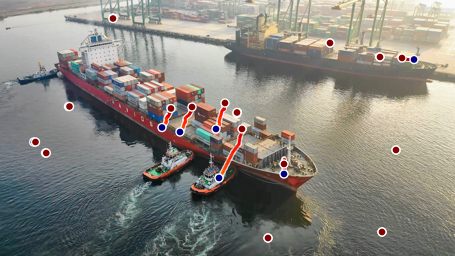}
    \end{subfigure}

    \begin{subfigure}[b]{0.3\textwidth}
        \centering
        \includegraphics[width=\linewidth]{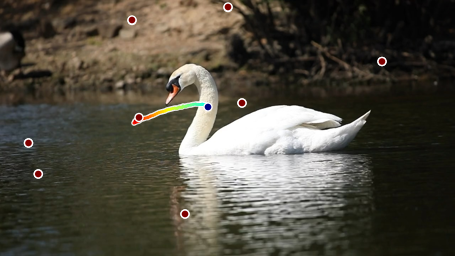}
    \end{subfigure}\hfill
    \begin{subfigure}[b]{0.3\textwidth}
        \centering
        \includegraphics[width=\linewidth]{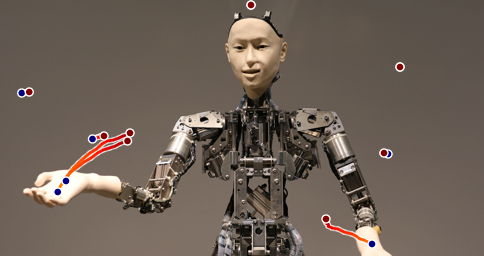}
    \end{subfigure}\hfill
    \begin{subfigure}[b]{0.3\textwidth}
        \centering
        \includegraphics[width=\linewidth]{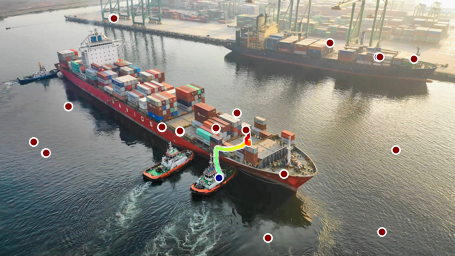}
    \end{subfigure}

    \begin{subfigure}[b]{0.3\textwidth}
        \centering
        \includegraphics[width=\linewidth]{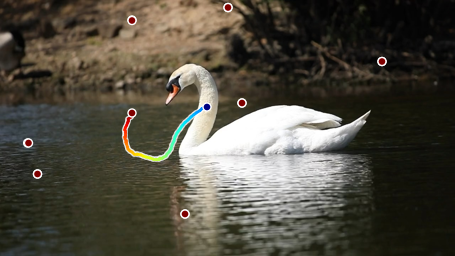}
    \end{subfigure}\hfill
    \begin{subfigure}[b]{0.3\textwidth}
        \centering
        \includegraphics[width=\linewidth]{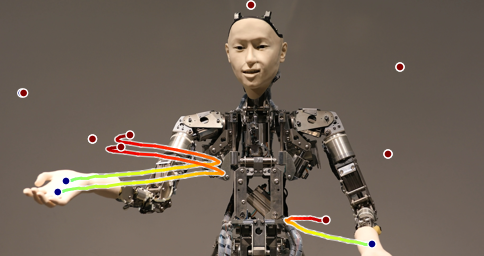}
    \end{subfigure}\hfill
    \begin{subfigure}[b]{0.3\textwidth}
        \centering
        \includegraphics[width=\linewidth]{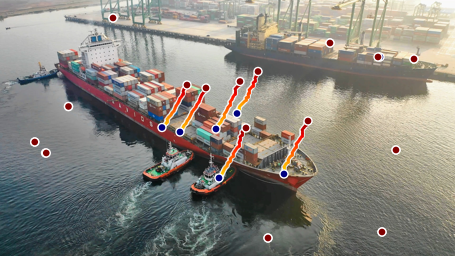}
    \end{subfigure}

    \caption{\textbf{Additional qualitative samples.} The first image in every column is the prompt, with arrows indicating pokes, i.e., where a specific start position should end. The remaining four images are samples from our models, illustrating trajectories of randomly selected query points. The color gradient of the trajectories acts as an indicator of how fast the motion is happening. The gradient starts at blue, moves to green, yellow, and finally ends at red.}
    \label{fig:app_qual_2}
\end{figure*}

We provide additional qualitative examples in \cref{fig:app_qual_1,fig:app_qual_2}.

